\documentclass[pdflatex,sn-mathphys-num]{sn-jnl}% Math and Physical Sciences Numbered Reference Style
\usepackage{graphicx}
\usepackage{multirow}
\usepackage{amsmath,amssymb,amsfonts}
\usepackage{amsthm}
\usepackage[title]{appendix}
\usepackage{xcolor}
\usepackage{textcomp}
\usepackage{manyfoot}
\usepackage{booktabs}
\usepackage{algorithm}
\usepackage{algorithmicx}
\usepackage{algpseudocode}

\usepackage{listings}
\usepackage{array}
\usepackage[caption=false,font=normalsize,labelfont=sf,textfont=sf]{subfig}
\usepackage{stfloats}
\usepackage{url}
\usepackage{verbatim}

\usepackage{hyperref}

\usepackage{newtxtext,newtxmath}  % 替代 \usepackage{times}

\usepackage{soul}
\usepackage[utf8]{inputenc}
\usepackage[switch]{lineno}
\usepackage{bbding}
\usepackage{adjustbox}
\usepackage{orcidlink}
\usepackage{fontawesome}
\usepackage{arydshln}
\usepackage{tabularray}

\usepackage{balance}

\theoremstyle{thmstyleone}%
\theoremstyle{thmstyletwo}%

\theoremstyle{thmstylethree}%

\begin{document}

\title[Article Title]{Progressive Bird’s-Eye-View Perception for Safety-Critical Autonomous Driving: A Comprehensive Survey}

\author[1,2]{\fnm{Yan} \sur{Gong}}\email{gongyan2020@foxmail.com}

\author[2]{\fnm{Naibang} \sur{Wang}}\email{7788521wnb@gmail.com}

\author[2]{\fnm{Jianli} \sur{Lu}}\email{lujianli364@163.com}

\author[2]{\fnm{Xinyu} \sur{Zhang}}\email{xyzhang@tsinghua.edu.cn}

\author*[1]{\fnm{Yongsheng} \sur{Gao}}\email{gaoys@hit.edu.cn}

\author[1]{\fnm{Jie} \sur{Zhao}}\email{jzhao@hit.edu.cn}

\author[2]{\fnm{Zifan} \sur{Huang}}\email{huangzf202507@163.com}

\author[2]{\fnm{Haozhi} \sur{Bai}}\email{haozhi0911@gmail.com}

\author[2]{\fnm{Nanxin} \sur{Zeng}}\email{nanceng077@gmail.com}

\author[2]{\fnm{Nayu} \sur{Su}}\email{su11031915@163,com}

\author[3]{\fnm{Lei} \sur{Yang}}\email{yangleils@outlook.com}

\author[4]{\fnm{Ziying} \sur{Song}}\email{songziying@bjtu.edu.cn}

\author[2]{\fnm{Xiaoxi} \sur{Hu}}\email{xiaoxihurail@gmail.com}

\author[2]{\fnm{Xinmin} \sur{Jiang}}\email{xinminj2001@gmail.com}

\author[5]{\fnm{Xiaojuan} \sur{Zhang}}\email{xzhang@i2r.a-star.edu.sg}

\author[6]{\fnm{Susanto} \sur{Rahardja}}\email{susantorahardja@ieee.org}

\affil*[1]{\orgdiv{State Key Laboratory of Robotics and System}, \orgname{Harbin Institute of Technology}, \orgaddress{\city{Harbin}, \postcode{150001}, \country{China}}}

\affil[2]{\orgdiv{State Key Laboratory of Intelligent Green Vehicle and Mobility, the School of Vehicle and Mobility}, \orgname{Tsinghua University}, \orgaddress{\city{Beijing}, \postcode{100084}, \country{China}}}

\affil[3]{\orgdiv{the School of Mechanical and Aerospace Engineering}, \orgname{Nanyang Technological University}, \orgaddress{ \country{Singapore}}}

\affil[4]{\orgdiv{Beijing Key Laboratory of Traffic Data Mining and Embodied Intelligence, School of Computer Science and Technology}, \orgname{Beijing Jiaotong University}, \orgaddress{\city{Beijing}, \country{China}}}

\affil[5]{\orgdiv{the Institute for Infocomm Research}, \orgname{A*STAR}, \orgaddress{ \country{Singapore}}}

\affil[6]{\orgdiv{the Engineering Cluster}, \orgname{the Singapore Institute of Technology}, \orgaddress{ \country{Singapore}}}

\abstract{Bird’s-Eye-View (BEV) perception has become a foundational paradigm in autonomous driving, enabling unified spatial representations that support robust multi-sensor fusion and multi-agent collaboration. As autonomous vehicles transition from controlled environments to real-world deployment, ensuring the safety and reliability of BEV perception in complex scenarios—such as occlusions, adverse weather, and dynamic traffic—remains a critical challenge. This survey provides the first comprehensive review of BEV perception from a safety-critical perspective, systematically analyzing state-of-the-art frameworks and implementation strategies across three progressive stages: single-modality vehicle-side, multimodal vehicle-side, and multi-agent collaborative perception. Furthermore, we examine public datasets encompassing vehicle-side, roadside, and collaborative settings, evaluating their relevance to safety and robustness. We also identify key open-world challenges—including open-set recognition, large-scale unlabeled data, sensor degradation, and inter-agent communication latency—and outline future research directions, such as integration with end-to-end autonomous driving systems, embodied intelligence, and large language models. An open-source repository summarizing existing methods and benchmark datasets is provided as a comprehensive resource for the community: \href{https://github.com/gongyan1/SafeBEV}{https://github.com/gongyan1/SafeBEV}.}

\keywords{Bird’s-Eye-View perception, autonomous driving, multimodal fusion, collaborative perception, safety, robustness}

\maketitle

\section{Introduction}\label{sec1}

With the rapid development of intelligent transportation, autonomous vehicles are moving from controlled environments to real-world deployment \cite{wang2024toward}, ~\cite{song2023graphalign++},~\cite{gong2024tclanenet},~\cite{zhang2023oblique},~\cite{liu2024glmdrivenet}. As the perceptual core, real-time and reliable environmental understanding is crucial for ensuring driving safety~\cite{Waymoopendataset},~\cite{li2020deep},~\cite{wang2025collaborative},~\cite{velasco2020autonomous},~\cite{zhang2023multi}. However, complex scenarios—e.g., illumination changes, adverse weather, dense traffic, and occlusions—can significantly degrade perception, affecting downstream decision-making and control. To enhance robustness, recent studies have explored multimodal sensor fusion~\cite{song2024robofusion},~\cite{lu2024lidar},~\cite{xiangV2IBEVFMultimodalFusion2023a},~\cite{singh2023vision},~\cite{li2023robust01},~\cite{gong2024steering},~\cite{zhao2023spatial}, and multi-agent collaborative perception~\cite{li2021learning},~\cite{wang2020v2vnet},~\cite{viana2021comparison}, which leverage complementary sensing and cross-agent communication. Nevertheless, aligning heterogeneous modalities and spatial representations across vehicle and infrastructure platforms remains a key challenge.

Notably, Bird’s-Eye-View (BEV) perception has emerged as a mainstream paradigm in autonomous driving~\cite{song2025graphbev}, providing a unified spatial representation that facilitates both multimodal fusion~\cite{man2023bev},~\cite{li2023delving} and multi-agent collaboration~\cite{zhao2024bev}. Given its central role, the progressive evolution of BEV-based perception—from single-modality to multimodality~\cite{feng2020deep} and further to collaborative frameworks—has become a key direction for enhancing the safety and robustness of autonomous driving systems~\cite{xu2022cobevt},~\cite{xiang2023v2i},~\cite{qiao2023cobevfusion},~\cite{dong2025data}. In the following sections, we systematically review BEV perception from three critical perspectives: \textit{\textbf{what}} BEV perception is, \textit{\textbf{why}} it is essential for autonomous driving safety, and \textit{\textbf{how}} it can be effectively implemented in multimodal and multi-agent scenarios.

\textit{\textbf{What}} – BEV perception serves as an efficient spatial representation paradigm that projects heterogeneous data from multiple sensor modalities, such as cameras~\cite{yang2023bevheight},~\cite{zou2023hft},~\cite{wang2019pseudo},~\cite{li2024bevformer}, LiDAR~\cite{sautier2024bevcontrast},~\cite{shi2019pointrcnn},~\cite{lang2019pointpillars}, and millimeter-wave radar~\cite{huang2024v2x},~\cite{palffy2020cnn},~\cite{zheng2022tj4dradset} into a unified BEV coordinate system. This projection constructs a consistent and structured spatial semantic map of the surrounding environment. By eliminating sensor-specific viewpoint variations, this top-down perspective facilitates the accurate perception and interpretation of spatial relationships between objects, significantly reducing the complexity of multi-view and multimodal data fusion~\cite{singh2023vision},~\cite{sun2023calico}.

\textit{\textbf{Why}} – Due to its unified and interpretable spatial representation~\cite{li2024bevformer}, BEV perception serves as an ideal foundation for multimodal fusion and multi-agent collaborative perception in autonomous driving~\cite{xu2022cobevt},~\cite{qiao2023cobevfusion},~\cite{huang2022bevdet4d},~\cite{lidelving}. By projecting heterogeneous sensor data—such as images, LiDAR, and radar—onto a common BEV plane, information from different modalities can be seamlessly aligned and integrated~\cite{li2024fast},~\cite{chang2023bev}. This unified coordinate system not only simplifies the fusion of vehicle-side and roadside sensors, but also enables efficient information sharing among multiple vehicles and infrastructure, thereby overcoming the limitations of single-vehicle perception. Furthermore, the structured and consistent semantics of BEV facilitate downstream tasks such as planning and control, making BEV a crucial bridge between perception and decision-making in complex, collaborative driving scenarios~\cite{wang2024bevgpt},~\cite{lu2024hierarchical},~\cite{jiang2024bevnav},~\cite{feng2025survey}.

\begin{figure*}[htbp]
    \centering
    \includegraphics[width=0.95\textwidth]{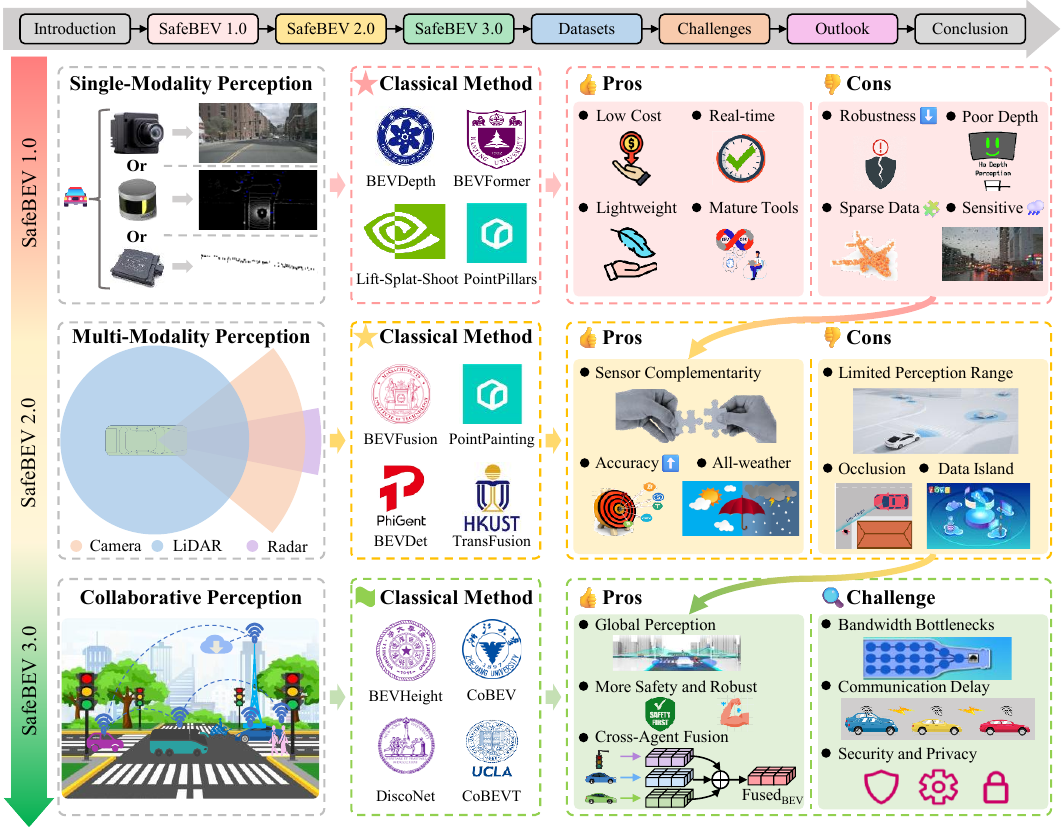} % Add your image path here
    \caption{Evolution of BEV perception frameworks in SafeBEV. The figure depicts the progressive development of BEV perception, advancing from single-modality vehicle-side perception (\textit{SafeBEV 1.0}), to multimodal vehicle-side perception (\textit{SafeBEV 2.0}), and finally to multi-agent collaborative perception (\textit{SafeBEV 3.0}). This evolution reflects a trajectory toward enhanced robustness, broader perceptual coverage, and improved collaboration. Each stage summarizes representative sensor configurations, canonical methodologies, as well as key advantages and challenges.}
    \label{framework_introduction}
\end{figure*}

\textit{\textbf{How}} – To address the challenges presented by increasingly complex and dynamic traffic scenarios, perception systems have continuously explored new paradigms to improve safety and robustness~\cite{huang2021bevdet},~\cite{sun2024robust},~\cite{liu2024h}. 
In this survey, we chart the progression of Safe BEV perception (SafeBEV) across three major stages, as illustrated in Fig.~\ref{framework_introduction}: \textbf{SafeBEV 1.0}—single-modality vehicle-side perception, \textbf{SafeBEV 2.0}—multimodality vehicle-side perception, and \textbf{SafeBEV 3.0}—multi-agent collaborative perception. The characteristics and advancements of each stage are elaborated in the following sections.

\textit{\textbf{SafeBEV 1.0: Single-modality vehicle-side perception.}} 
This stage employs a single sensor (e.g., camera or LiDAR) for BEV-based scene understanding. Early camera-based methods relied on homography~\cite{bertozz1998stereo},~\cite{gong2023feature},~\cite{gong2025skipcrossnets}, which lacked robustness in complex scenes. Recent approaches shift toward data-driven BEV modeling, broadly categorized into sparse and dense paradigms. Sparse methods~\cite{wang2019pseudo},~\cite{philion2020lift},~\cite{hu2021fiery} estimate depth to lift 2D features or generate point clouds, then voxelize them into BEV, but are sensitive to depth quality. Dense methods~\cite{li2023delving},~\cite{roddick2020predicting},~\cite{yang2021projecting},~\cite{saha2022translating},~\cite{chen2022graph} adopt MLPs or transformers (e.g., BEVFormer~\cite{li2024bevformer}, PETR~\cite{liu2022petr}) to directly map 2D features to BEV via nonlinear projection or cross-view attention. Hybrid designs such as BEVDepth~\cite{li2023bevdepth} and BEVDet~\cite{huang2021bevdet} integrate depth prediction to enhance dense BEV modeling. For LiDAR, perception pipelines~\cite{zhou2023fastpillars},~\cite{zhang2024safdnet},~\cite{qiu2025pc} typically voxelize point clouds or apply SparseConv~\cite{shi2020pv} and PointNet~\cite{qi2017pointnet} for BEV feature extraction, balancing spatial resolution and computational efficiency~\cite{zhao2024bev}.

\textit{\textbf{SafeBEV 2.0: Multimodality Vehicle-Side Perception.}}
This stage advances BEV perception by integrating heterogeneous sensors—cameras, LiDAR, and radar—to overcome the limitations of single-modality systems and enhance robustness under occlusion and adverse weather~\cite{thakur2024depth}. Recent efforts span five major fusion paradigms: camera-radar~~\cite{kim2023craft},~\cite{yu2023sparsefusion3d},~\cite{schramm2024bevcar}, camera-LiDAR~\cite{liu2023bevfusion},~\cite{liang2022bevfusion},~\cite{hu2023fusionformer}, radar-LiDAR~\cite{yang2020radarnet},~\cite{wang2022interfusion}, camera-LiDAR-radar~\cite{man2023bev},~\cite{malawade2022hydrafusion},~\cite{chen2023futr3d}, and temporal fusion~\cite{huang2022bevdet4d},~\cite{wu2020motionnet},~\cite{zhang2022beverse},~\cite{li2023towards}.
Each modality combination leverages complementary properties: camera-based semantics with geometry-aware LiDAR/radar for accuracy and reliability; LiDAR’s spatial precision with radar’s velocity sensing for long-range, all-weather perception; and full tri-modal fusion for comprehensive BEV representations. Temporal fusion further promotes temporal consistency in dynamic scenes. These strategies collectively strengthen the safety, adaptability, and reliability of BEV perception in real-world autonomous driving~\cite{alaba2024emerging},~\cite{yu2025samfusion3d},~\cite{xiong2023lxl}.

\textit{\textbf{SafeBEV 3.0: Multi-agent Collaborative Perception.}}
With the advancement of vehicle-to-everything (V2X) technologies, autonomous vehicles can exchange and jointly infer perception information across vehicles and infrastructure, mitigating the limitations of single-agent perception~\cite{zhang2024v2x},~\cite{liu2024v2x},~\cite{yinV2VFormerMultiModalVehicletoVehicle2024}. By aggregating multi-source sensor data within a unified BEV space, collaborative perception enables holistic environmental modeling crucial for safe navigation in dynamic traffic. Representative frameworks such as V2VNet~\cite{wangV2VNetVehicletoVehicleCommunication2020}, DiscoNet~\cite{li2021learning}, and CoBEVT~\cite{xu2022cobevt} employ feature compression, bandwidth-efficient protocols, and distributed inference to support real-time, scalable collaboration with minimal communication cost. Furthermore, spatiotemporal fusion of multi-agent observations~\cite{xiangV2IBEVFMultimodalFusion2023a},~\cite{shi2023mcot},~\cite{yuVehicleInfrastructureCooperative3D2023},~\cite{rossleUnlockingInformationTemporal2024},~\cite{li2024coformernet},~\cite{wei2023asynchrony} enhances global situational awareness and improves the perception of occluded or distant targets. This paradigm also underpins higher-level functionalities such as collective decision-making, collaborative trajectory planning, and multi-agent control~\cite{han2023collaborative}, ~\cite{bai2024survey}, marking a critical step toward large-scale, safe, and intelligent autonomous driving.

Given the rapid progress in BEV perception, several surveys have emerged: Ma et al.~\cite{ma2024vision} focus on camera-only methods, summarizing three decades of development; Li et al.~\cite{li2023delving} review monocular and multimodal BEV perception at the vehicle level; Zhao et al.~\cite{zhao2024bev} categorize vehicle-to-vehicle (V2V) collaboration paradigms; and Singh et al.~\cite{singh2023vision} examine vision–radar fusion under adverse conditions. While these works offer valuable overviews from perspectives such as camera-only, multimodality fusion, and V2V collaboration, key gaps remain: (1) limited analysis of BEV perception from safety and robustness perspectives; (2) insufficient coverage of roadside BEV systems and broader collaboration paradigms, including vehicle-to-infrastructure (V2I) and infrastructure-to-infrastructure (I2I); and (3) a lack of in-depth discussion on large-scale vehicle-side and multi-agent datasets. Moreover, the integration of BEV perception with emerging paradigms—such as end-to-end (E2E) autonomous driving, large language models (LLMs), and embodied intelligence—remains underexplored.

This paper presents the first comprehensive survey of vehicle-side and collaborative BEV perception methods from the perspective of safety and robustness. We categorize existing approaches into three stages: \textit{\textbf{single-modality vehicle-side perception}}, \textit{\textbf{multimodality vehicle-side perception}}, and \textit{\textbf{multi-agent collaborative perception}}. In addition, we systematically review key public datasets, evaluate their support for safety and robustness, and establish corresponding metrics and benchmarks to guide future research. To promote accessibility and reproducibility, we maintain a GitHub repository with method implementations, dataset guidelines, and example code. Finally, we identify critical safety challenges—such as open-world deployment, large-scale unlabeled data, sensor degradation, and multi-agent communication delay—and discuss future trends including integration with embodied intelligence, end-to-end autonomous driving, and large language models. The main contributions of this survey are:

\begin{itemize}
\item To our knowledge, this is the first work to systematically define BEV perception systems in three stages—single-modality, multimodality, and multi-agent collaboration—from the perspective of safety and robustness, and to comprehensively analyze their respective strengths and limitations across diverse scenarios.

\item We systematically review public datasets for vehicle-side and collaborative BEV perception, assess their support for safety and robustness, and establish corresponding evaluation benchmarks. Additionally, we provide an open-source guide detailing dataset characteristics, selection criteria, and benchmark results for various methods.

\item We thoroughly examine critical security challenges for BEV perception in open-world settings—such as open-set recognition, large-scale unlabeled data, sensor degradation, and multi-agent communication delay—and discuss its integration with end-to-end autonomous driving, embodied intelligence, and large language models.
\end{itemize}

The remainder of this paper is organized as follows. Section~\ref{sec2} reviews BEV perception using single-modality vehicle-side sensors, followed by Section~\ref{sec3}, which explores multimodal fusion strategies in terms of safety and robustness. Section~\ref{sec4} examines multi-agent collaborative BEV perception, including roadside sensing and V2X collaboration. Section~\ref{sec5} surveys public datasets across vehicle-side, roadside, and collaborative settings, highlighting current limitations and future directions. Finally, Sections~\ref{sec6} and~\ref{sec7} summarize challenges and outline future research toward safe and robust BEV perception for real-world autonomy.

\section{SafeBEV 1.0: Vehicle-based Single-Modality BEV perception method}\label{sec2}

Single-modality vehicle-side BEV perception serves as a foundational stage in autonomous driving, offering efficient top-down scene understanding using either camera or LiDAR sensors. While this approach benefits from reduced system complexity and computational cost, its robustness remains limited under adverse conditions. Camera-based methods are sensitive to illumination changes, occlusions, and depth estimation errors, whereas LiDAR-based methods face challenges from data sparsity and weather-induced degradation. This section reviews the strengths and limitations of single-modality BEV perception and categorizes existing methods into camera-based (Section~\ref{II-A}) and LiDAR-based (Section~\ref{II-B}) approaches.

\begin{figure*}[t]
    \centering
    \includegraphics[width=\textwidth]{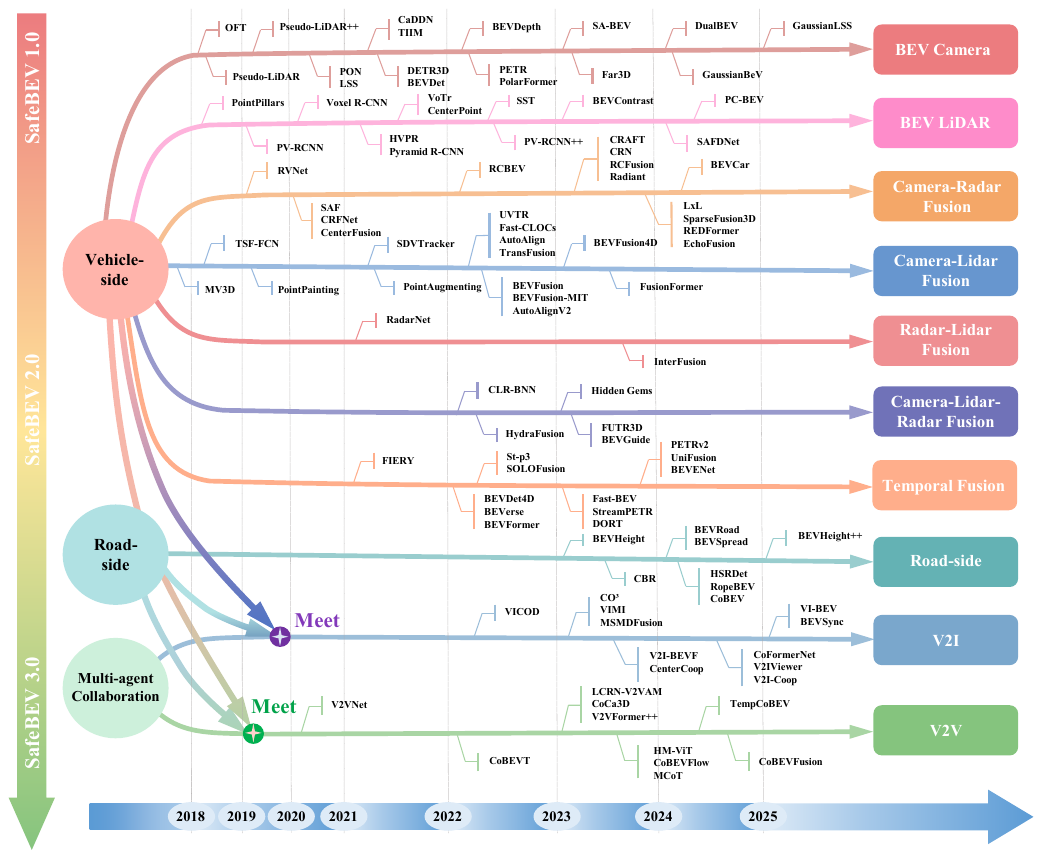}
    \caption{Evolution of BEV perception methods (2018–2025) under the SafeBEV paradigm. The timeline is divided into three stages: SafeBEV 1.0 (single-modality, vehicle-side perception), SafeBEV 2.0 (multimodal vehicle-side perception), and SafeBEV 3.0 (multi-agent collaborative perception with roadside integration). Methods are categorized by sensor configurations and deployment locations, including BEV camera, BEV LiDAR, camera–radar fusion, camera–LiDAR fusion, radar–LiDAR fusion, camera–LiDAR–radar fusion, roadside perception, and collaborative paradigms (e.g., V2I and V2V). “Meet” points indicate convergence or fusion across different branches.}
    \label{timeline_bev}
\end{figure*}

\subsection{BEV Camera}\label{II-A}
\begin{figure*}[t]
    \centering
    \includegraphics[width=0.9\textwidth]{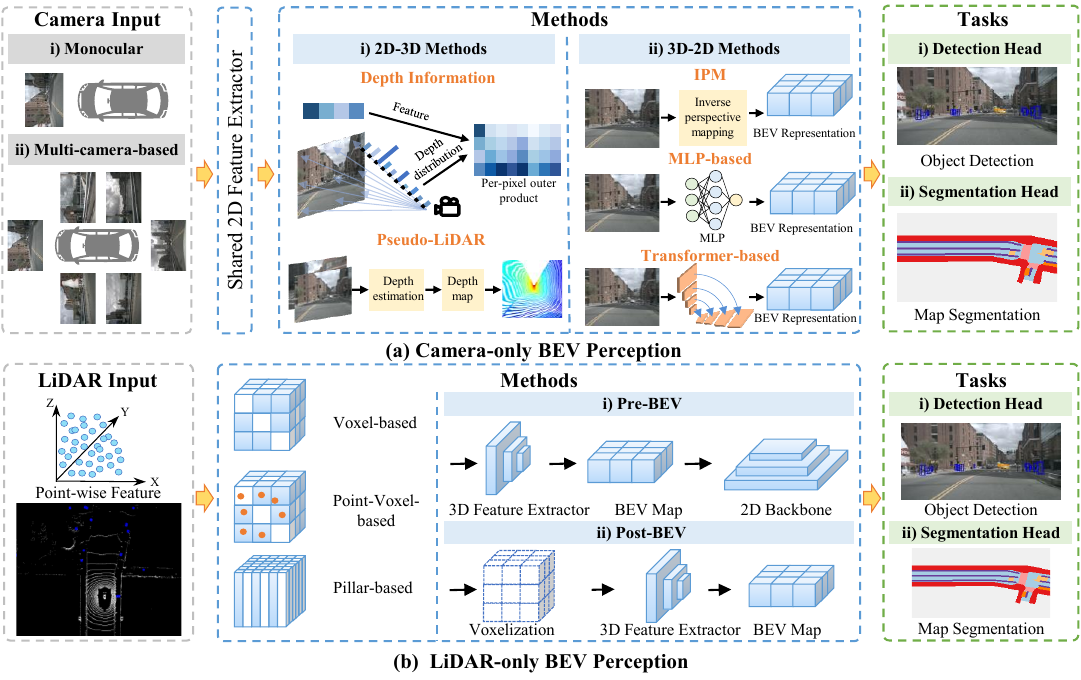}
    \caption{The overall framework of SafeBEV 1.0. This figure summarizes the typical pipelines for BEV perception using either (a) camera-only or (b) LiDAR-only inputs. For camera-based BEV perception (top), monocular or multi-camera inputs are first encoded into perspective-view features and then transformed into BEV space via 2D–3D or 3D–2D methods, including Pseudo-LiDAR, IPM-based, MLP-based, and Transformer-based frameworks. For LiDAR-based BEV perception (bottom), voxelization or point-voxel operations are applied to raw point clouds, followed by BEV map construction either before or after 3D backbone processing (i.e., pre-BEV or post-BEV strategies).}
    \label{pipeline}
\end{figure*}

Vehicle-side camera-only BEV perception faces the fundamental challenge of inferring spatially consistent BEV representations from 2D images~\cite{tao2023pseudo}. Recent advances~\cite{wang2019pseudo},~\cite{you2019pseudo},~\cite{lu2025gaussianlss}, illustrated in Fig.~\ref{timeline_bev}, typically adopt a two-stage pipeline: extracting features from perspective-view images, followed by spatial projection into the BEV plane for downstream tasks (see Fig.~\ref{pipeline} (a)). Table~\ref{Single-Modality BEV Perception} categorizes representative methods by camera configuration (monocular vs. multi-camera) and feature transformation strategy—either \textit{3D-to-2D}, which infers intermediate 3D geometry prior to projection, or \textit{2D-to-3D}, which directly learns spatial reasoning through BEV mapping. This section reviews key approaches in each category, focusing on their core assumptions, encoding mechanisms, and spatial transformation designs.

\subsubsection{Monocular}
  
\paragraph{\textbf{2D-to-3D Methods}} 
To resolve the inherent depth ambiguity in monocular BEV perception, existing methods typically adopt a two-stage strategy: first extracting 2D features from perspective-view images, followed by lifting them into 3D space through depth estimation to enable accurate projection into the BEV plane.

\textit{``Pseudo-LiDAR'' methods.} These methods convert monocular or stereo imagery into depth maps, which are back-projected into 3D space to simulate LiDAR point clouds~\cite{wang2019pseudo}. AM3D~\cite{ma2019accurate} extends this paradigm by integrating a monocular 3D detection framework with feature fusion to improve detection accuracy. Simonelli et al.~\cite{simonelli2021we} further enhance reliability through a 3D confidence estimation module.

\textit{Depth distribution methods.} These approaches model per-pixel depth as probability distributions and directly project them into the BEV space, enabling efficient geometric reasoning without explicit point cloud construction. OFT~\cite{roddick2018orthographic} addresses perspective distortion by orthographically projecting image features into BEV and aggregating them along the height axis. CaDDN~\cite{reading2021categorical} builds upon this by predicting categorical depth distributions and leveraging LiDAR-supervised projection geometry to enhance BEV feature representation from monocular input.

\paragraph{\textbf{3D-to-2D Methods}}
In contrast to 2D-to-3D paradigms that depend on depth estimation, 3D-to-2D methods leverage predefined 3D priors to guide the projection of image-plane features into the BEV space.

\textit{IPM-based methods.}
A class of methods leverages inverse perspective mapping (IPM) to incorporate geometric priors for transforming features from the image plane to the BEV domain. Early studies tackled the inherent depth ambiguity in monocular inputs by utilizing camera imaging principles~\cite{mallot1991inverse}, enabling the projection of perspective-view segmentations into BEV space. Building on this foundation, Kim et al.~\cite{kim2019deep} integrated rectified front-view images with IPM and employed CNNs for object detection directly in BEV space, establishing a foundational framework for IPM-guided BEV perception.

\textit{MLP-based methods.}
Another research line explores multilayer perceptron (MLP) architectures to learn implicit mappings from perspective views to BEV representations in a data-driven manner, without explicit geometric modeling. PYVA~\cite{yang2021projecting} introduces a recurrent MLP framework to enhance spatial consistency, complemented by a transformer-based module for cross-view alignment. BEV-LaneDet~\cite{wang2022bev} further improves geometric normalization and multi-scale fusion via a virtual camera design and an FPN-inspired feature aggregation strategy.

\textit{Transformer-based methods.}
A more recent paradigm employs transformer architectures to perform view transformation through attention-based mechanisms, typically using BEV queries and decoder modules. PON~\cite{roddick2020predicting} initiates this direction by projecting epipolar features into BEV for semantic segmentation. STSU~\cite{can2021structured} advances this with structured queries in a unified Transformer for joint road topology estimation and object detection. To facilitate fine-grained spatial reasoning, TIIM~\cite{saha2022translating} models the transformation as a sequence-to-sequence process with dual attention mechanisms. PanopticBEV~\cite{gosala2022bird} disentangles vertical and planar region projections using two specialized Transformers, while HFT~\cite{zou2023hft} fuses camera-aware and camera-agnostic pathways to enhance BEV feature representation.

\subsubsection{Multi-camera}

\paragraph{\textbf{2D-to-3D Methods}} 
In multi-camera setups, 2D-to-3D methods follow similar principles to monocular systems. However, 2D-to-3D methods in multi-camera systems offer enhanced spatial reasoning and denser scene reconstruction compared to monocular setups.

\textit{``Pseudo-LiDAR'' methods.}
Stereo-based approaches exploit denser depth cues to generate pseudo-LiDAR point clouds, offering finer spatial details than monocular estimation. Nevertheless, their constrained field of view limits global perception and motivates the development of range-aware enhancements~\cite{zhang2021pseudo}. Pseudo-LiDAR++~\cite{you2019pseudo} refines early designs using improved stereo depth estimation, while E2E Pseudo-LiDAD~\cite{qian2020end} introduces a differentiable coordinate transformation enabling end-to-end training. Subsequent advances incorporate confidence-aware depth refinement~\cite{meng2022accurate} and efficient stereo matching~\cite{li2022real} to further boost accuracy and runtime efficiency.

\textit{Depth distribution methods.}
Another prevalent strategy relies on estimating per-pixel depth distributions and transforming image features into 3D space. LSS~\cite{philion2020lift} estimates probabilistic depth distributions and projects image features into 3D volumes via outer product operations, generating unified BEV feature maps. Building on this paradigm, works such as BEVDet~\cite{huang2021bevdet}, BEVDepth~\cite{li2023bevdepth}, and MatrixVT~\cite{zhou2023matrixvt} adopt LSS-style view transformation and introduce architectural refinements to mitigate the original framework’s limitations in accuracy, efficiency, and scalability.

\paragraph{\textbf{3D-to-2D Methods}} 
The effectiveness of 3D-to-2D approaches in monocular setups has motivated their adaptation to multi-camera systems, where multiple calibrated views enhance spatial context modeling and reduce occlusion effects.

\textit{MLP-based methods.}
These approaches offer a lightweight and effective means for mapping multi-view image features into the BEV domain. NEAT~\cite{chitta2021neat} employs an implicit decoder based on MLPs to fuse features for joint BEV segmentation and motion planning. VPN~\cite{pan2020cross} introduces an MLP-based visual relation module to aggregate multi-scale contextual features across views. To improve robustness against calibration noise, Sun et al.~\cite{sun2024robust} propose a dual-space positional encoding scheme coupled with MLP fusion for accurate BEV map prediction.

\textit{Transformer-based methods.}
Transformers demonstrate strong potential for multi-camera BEV perception through implicit depth modeling and global feature aggregation. LaRa~\cite{bartoccioni2023lara} incorporates geometry-aware ray embeddings and cross-attention to enhance spatial reasoning. PETR~\cite{liu2022petr} addresses the projection artifacts of DETR3D~\cite{wang2022detr3d} by eliminating explicit view transformations via an end-to-end architecture. PolarFormer~\cite{jiang2023polarformer}, PolarDETR~\cite{chen2022polar}, and Ego3RT~\cite{lu2022learning} enrich BEV feature encoding with polar coordinate priors. Graph-DETR3D~\cite{chen2022graph} leverages a dynamic graph-based mechanism to associate object queries with informative image regions, mitigating depth uncertainty. CoBEVT~\cite{xu2022cobevt} employs a collaborative Transformer design for efficient multi-view BEV segmentation.

\subsubsection{Accuracy–Robustness–Safety Analysis}
The performance of camera-only BEV perception depends heavily on input quality and algorithmic resilience. 2D-to-3D methods improve spatial awareness by estimating depth and lifting features into 3D space, but their reliability degrades under adverse conditions such as low light, occlusion, or inclement weather. In contrast, 3D-to-2D methods bypass explicit depth prediction by incorporating structural priors directly into the BEV representation, though they may compromise geometric fidelity. Achieving a balance between accurate 3D reconstruction and efficient BEV representation remains a central challenge in vision-only BEV perception systems.

\begin{table*}[!t]
\centering
\setlength{\tabcolsep}{7pt} 
\caption{Summary of representative vehicle-side BEV perception methods categorized by modality, task, and dataset. ``Type'' denotes the sensor type: ``C'' for camera and ``L'' for LiDAR. ``Modality'' indicates camera configuration: ``SC'' for single camera and ``MC'' for multi-camera setups. ``Task'' includes object detection (OD) and map semantic segmentation (MapSeg). ``Dataset'' abbreviations: KITTI, nuScenes, Waymo Open Dataset (WOD), Lyft, and Argoverse. ``-'' indicates that the corresponding field is not specified.}
\label{Single-Modality BEV Perception}
\resizebox{\linewidth}{!}{
\begin{tabular}{clccc*{5}{c}c}
% \hline
\toprule
\multirow{2}{*}{\textbf{Type}} & 
\multirow{2}{*}{\textbf{Method}} & 
\multirow{2}{*}{\textbf{Venue}} & 
\multirow{2}{*}{\textbf{Camera Setup}} & 
\multirow{2}{*}{\textbf{Task}} & 
\multicolumn{5}{c}{ \rule{0pt}{1.1em}\textbf{Datasets}} &
\multirow{2}{*}{\textbf{Code}} \\
% \cline{6-10}
\cmidrule(lr){6-10} 
 & & & & & \rule{0pt}{1.0em}\textbf{KITTI} & \textbf{NuScenes} & \textbf{WOD} & \textbf{Lyft} & \textbf{Argoverse} & \\
\midrule
\multirow{19}{*}{{C}}
& Pseudo-LiDAR~\cite{wang2019pseudo} & CVPR 2019  & SC & OD  & \Checkmark &   &   &   &   & \href{https://github.com/mileyan/pseudo_lidar}{\faGithub} \\
& PON~\cite{roddick2020predicting}& arXiv 2020 & SC & MapSeg &   & \Checkmark &   &   & \Checkmark & \href{https://github.com/tom-roddick/mono-semantic-maps}{\faGithub} \\
& CaDDN~\cite{reading2021categorical} & CVPR 2021  & SC  & OD  & \Checkmark &   &   &   &   & \href{https://github.com/TRAILab/CaDDN}{\faGithub} \\
& TIIM~\cite{saha2022translating}  & ICRA 2022 & SC & Seg  &   & \Checkmark &   &   & \Checkmark & \href{https://github.com/avishkarsaha/translating-images-into-maps}{\faGithub} \\
& LSS~\cite{philion2020lift}  & ECCV 2020  & MC  & Seg  &  & \Checkmark  &   & \Checkmark  &   & \href{https://github.com/nv-tlabs/lift-splat-shoot}{\faGithub} \\
& Pseudo-LiDAR++ \cite{you2019pseudo}  & ICLR 2020  & MC  & OD  & \Checkmark &   &   &   &   & \href{https://github.com/mileyan/Pseudo_Lidar_V2}{\faGithub} \\
& BEVDet~\cite{huang2021bevdet}  & arXiv 2021  & MC  & OD  &  & \Checkmark  &   &   &   & \href{https://github.com/HuangJunJie2017/BEVDet}{\faGithub} \\
& LaRa~\cite{bartoccioni2023lara} & CoRL 2022 & MC    & Seg     &   & \Checkmark &   &   &   & \href{https://github.com/valeoai/LaRa}{\faGithub} \\
& DETR3D~\cite{wang2022detr3d} & CoRL 2021 & MC    & OD     &   & \Checkmark &   &   &   & \href{https://github.com/WangYueFt/detr3d}{\faGithub} \\
& PETR~\cite{liu2022petr} & ECCV 2022 & MC    & OD     &   & \Checkmark &   &   &   & \href{https://github.com/megvii-research/PETR}{\faGithub} \\
& BEVDepth~\cite{li2023bevdepth} & AAAI 2023 & MC    & OD     &   & \Checkmark &   &   &   & \href{https://github.com/Megvii-BaseDetection/BEVDepth}{\faGithub} \\
& PolarFormer~\cite{jiang2023polarformer} & AAAI 2023 & MC & OD &   & \Checkmark &   &   &   & \href{https://github.com/fudan-zvg/PolarFormer}{\faGithub} \\
& SA-BEV~\cite{zhang2023sa} & ICCV 2023  & MC    & OD &   & \Checkmark &   &   &   & \href{https://github.com/mengtan00/SA-BEV}{\faGithub} \\
& Far3D~\cite{jiang2024far3d} & AAAI 2024  & MC    & OD &   &   &   &   & \Checkmark & \href{https://github.com/megvii-research/Far3D}{\faGithub} \\
& DualBEV~\cite{li2024dualbev}  & ECCV 2024  & MC  & OD  &   & \Checkmark &   &   &   & \href{https://github.com/PeidongLi/DualBEV}{\faGithub} \\
& DA-BEV~\cite{jiang2024bev} & ECCV 2024  & MC    & OD/MapSeg &  &\Checkmark&  & \Checkmark &   & \href{https://github.com/xdjiangkai/DA-BEV}{\faGithub} \\
& GaussianBeV~\cite{chabot2025gaussianbev} & WACV 2025 & MC & MapSeg &   & \Checkmark &   &   &   & -- \\
& GaussianLSS~\cite{lu2025gaussianlss} & arXiv 2025 & MC & OD/MapSeg &   & \Checkmark &   &   &   & \href{https://github.com/HCIS-Lab/GaussianLSS}{\faGithub} \\
\midrule
% \multirow{14}{*}{BEV LiDAR} 
\multirow{15}{*}{{L}}
& PointPillars~\cite{lang2019pointpillars} & CVPR 2019  & - & OD & \Checkmark &   &   &   &   & \href{https://github.com/zhulf0804/PointPillars}{\faGithub} \\
& PV-RCNN~\cite{shi2020pv} & CVPR 2020  & -     & OD & \Checkmark &   & \Checkmark &   &   & \href{https://github.com/sshaoshuai/PV-RCNN}{\faGithub} \\
& Voxel R-CNN~\cite{deng2021voxel} & AAAI 2021  & - & OD & \Checkmark &   & \Checkmark &   &   & \href{https://github.com/djiajunustc/Voxel-R-CNN}{\faGithub} \\
& CenterPoint~\cite{yin2021center} & CVPR 2021  & - & OD &  & \Checkmark  & \Checkmark &   &   & \href{https://github.com/tianweiy/CenterPoint}{\faGithub} \\
& VoTr~\cite{mao2021voxel} & ICCV 2021  & -     & OD  & \Checkmark &   & \Checkmark &   &   & \href{https://github.com/PointsCoder/VOTR}{\faGithub} \\
& HVPR~\cite{noh2021hvpr} & CVPR 2021  & -     & OD  & \Checkmark &   &   &   &   & \href{https://github.com/cvlab-yonsei/HVPR}{\faGithub} \\
& Pyramid R-CNN~\cite{mao2021pyramid} & ICCV 2021  & - & OD & \Checkmark &   & \Checkmark &   &   & \href{https://github.com/PointsCoder/Pyramid-RCNN}{\faGithub} \\
& SST~\cite{fan2022embracing} & CVPR 2022  & -     & OD &   &   & \Checkmark &   &   & \href{https://github.com/catherine-lisa/SST_multisweeps}{\faGithub} \\
& PillarNet~\cite{shi2022pillarnet} & ECCV 2022  & - & OD &   & \Checkmark & \Checkmark &   &   & \href{https://github.com/VISION-SJTU/PillarNet}{\faGithub} \\
& AFDetV2~\cite{hu2022afdetv2} & AAAI 2022  & -     & OD &   &   & \Checkmark &   &   &-- \\
& PV-RCNN++~ \cite{shi2023pv} & IJCV 2023  & -     & OD  &   &   & \Checkmark &   &   &-- \\
& FastPillars~\cite{zhou2023fastpillars} & AAAI 2024 & - & OD &   & \Checkmark & \Checkmark &   &   & \href{https://github.com/StiphyJay/FastPillars}{\faGithub} \\
& SAFDNet~\cite{zhang2024safdnet} & CVPR 2024 & -  & OD  &   & \Checkmark & \Checkmark &   & \Checkmark & \href{https://github.com/zhanggang001/HEDNet}{\faGithub} \\
& BEVContrast~\cite{sautier2024bevcontrast}  & 3DV 2024 & -  & MapSeg & \Checkmark & \Checkmark &   &   &   & \href{https://github.com/valeoai/BEVContrast}{\faGithub} \\
& PC-BEV~\cite{qiu2025pc} &  AAAI 2025  & -         & MapSeg  & \Checkmark & \Checkmark &   &   &   & \href{https://github.com/skyshoumeng/PC-BEV}{\faGithub} \\
\bottomrule
\end{tabular}
}
\vspace{1mm}
\end{table*}

\subsection{BEV LiDAR}\label{II-B}
LiDAR-based BEV methods focus on object detection by generating BEV feature maps via feature extraction and view transformation~\cite{yin2021center},~\cite{hu2022afdetv2},~\cite{he2020structure},~\cite{liang2020rangercnn}, as shown in Table~\ref{Single-Modality BEV Perception}. LiDAR offers richer depth cues than cameras but suffers from data sparsity. Methods are typically categorized as Pre-BEV (feature extraction before projection) or Post-BEV (feature extraction after projection).

\subsubsection{Pre-BEV Methods}
Pre-BEV methods are commonly divided into voxel-based and point-voxel hybrid approaches. Voxel-based methods discretize point clouds into regular grids to enable structured processing, whereas point-voxel hybrids aim to combine the geometric precision of point-based methods with the computational efficiency of voxel representations.

\paragraph{\textbf{Voxel-based Methods}} 
Voxel-based methods convert irregular point clouds into dense voxel grids to enable 3D convolutional processing. SECOND~\cite{yan2018second} improves upon VoxelNet by introducing sparse convolution to reduce computational overhead. Voxel R-CNN~\cite{deng2021voxel} adopts a two-stage framework that maintains competitive accuracy with reduced cost. InfoFocus~\cite{wang2020infofocus} introduces density-aware enhancement to refine coarse voxel features under sparsity. Transformer-based variants, such as VoTr~\cite{mao2021voxel} and SST~\cite{fan2022embracing}, integrate attention mechanisms with sparse convolutions to improve feature extraction and scalability.

\paragraph{\textbf{Point-voxel-based Methods}} 
These methods integrate the advantages of point- and voxel-based features, as demonstrated by PV-RCNN~\cite{shi2020pv}, PVGNet~\cite{miao2021pvgnet}, and PV-RCNN++~\cite{shi2023pv}, achieving improved accuracy, lower computational cost, and faster inference through effective fusion strategies. HVPR~\cite{noh2021hvpr} introduces an attention-based multi-scale feature module to extract scale-aware representations, alleviating the inherent sparsity and irregularity of point clouds. Pyramid R-CNN~\cite{mao2021pyramid} further addresses these challenges by incorporating a RoI grid pyramid, grid-based attention, and a density-aware radius prediction module.

\subsubsection{Post-BEV Methods}
Traditional LiDAR perception pipelines often rely on voxel-based representations processed by 3D or sparse 3D convolutions, which are computationally intensive and challenging to deploy in real-time industrial systems~\cite{li2023delving}, ~\cite{aksoy2020salsanet}, ~\cite{zhou2022centerformer}. Post-BEV methods address this limitation by shifting feature extraction to the BEV space, thereby avoiding 3D convolutions altogether.

Early approaches such as RT3D~\cite{zeng2018rt3d} and PIXOR~\cite{yang2018pixor} project 3D point clouds onto 2D grids or image-like representations to enable efficient object detection. To address feature distortion caused by varying distances and resolutions, BirdNet~\cite{beltran2018birdnet} proposes a scale-invariant encoding scheme. Methods like PointPillars~\cite{lang2019pointpillars} and PillarNet~\cite{shi2022pillarnet} enhance BEV representations by applying 2D convolutions to pseudo-image features. PolarNet~\cite{zhang2020polarnet} further models radial spatial dependencies using polar coordinate systems. Recent advances employ attention-based encoder-decoder architectures, such as SalsaNet~\cite{aksoy2020salsanet} and CenterFormer~\cite{zhou2022centerformer}, to improve detection robustness in complex driving scenarios.

\subsubsection{Accuracy–Robustness–Safety Analysis}
Under adverse weather (e.g., rain, fog, dust), LiDAR-only systems degrade due to signal scattering and absorption, causing increased noise, reduced range, and incomplete point returns. These effects aggravate point cloud sparsity and occlusion, heightening the risk of missing critical targets such as pedestrians or small obstacles. Addressing this requires feature extraction mechanisms robust to data incompleteness and sensor degradation.

\subsection{Challenges and Limitations}\label{II-C}
Single-modality BEV perception with cameras or LiDAR enables cost-effective detection and mapping but struggles under complex conditions. Camera-based methods degrade in low light and adverse weather, while LiDAR, despite accurate depth, suffers from sparsity, noise, and hardware vulnerability—leading to missed distant or low-reflectivity objects. Additionally, perspective transformation hampers 3D structure recovery, limiting height estimation and spatial reasoning under occlusion or uneven terrain. To address these limitations, recent works~\cite{wang2024unibev},~\cite{yin2024fusion},~\cite{hao2025mapfusion} advocate multimodal fusion, integrating visual and depth cues to enhance perception accuracy, robustness, and safety.

\section{SafeBEV 2.0: Vehicle-based Multimodality BEV perception method}\label{sec3}
Vehicle-side multimodal BEV perception enhances environmental understanding by integrating complementary sensor data, overcoming the limitations of single-modality systems and improving robustness and safety. This section categorizes fusion strategies into five types: camera–radar, camera–LiDAR, radar–LiDAR, camera–LiDAR–radar, and temporal fusion, as summarized in Table~\ref{Multi-Modality BEV Perception}. Each is further classified by fusion stage: single-stage methods fuse data at one point for efficiency, while multi-stage methods allow iterative cross-modal interaction for improved alignment and synergy.

\begin{table*}[!t]
\centering
% \small
\setlength{\tabcolsep}{6pt} 
\caption{Overview of representative vehicle-side BEV fusion perception methods categorized by sensor combinations, fusion stages, tasks, and datasets. ``Type'' indicates sensor modalities: ``C'' for Camera, ``L'' for LiDAR, and ``R'' for Radar. ``Modality'' includes ``SSF'' (Single-Stage Fusion) and ``MSF'' (Multi-Stage Fusion). ``Task'' refers to the perception objective, such as Object Detection (OD), Map Semantic Segmentation (MapSeg), and Multi-Object Tracking (MOT). ``--'' indicates that the field is unspecified or not publicly available.}
\label{Multi-Modality BEV Perception}
\resizebox{1.0\textwidth}{!}{
\begin{tabular}{m{2.5cm}<{\centering}m{3.5cm}<{\raggedright}m{2.2cm}<{\centering}m{2.2cm}<{\centering}m{1.5cm}<{\centering}m{4.5cm}<{\centering}m{1.0cm}<{\centering}}
\toprule
Type & Method & Venue & Fusion Strategy & Task & Datasets & Code\\
\midrule
\multirow{12}{*}{C\&R}
& CRAFT \cite{kim2023craft}  & AAAI 2023  & SSF    & OD & NuScenes & -- \\
& CRN \cite{kim2023crn}   & ICCV 2023  & SSF       & OD & NuScenes & \href{https://github.com/youngskkim/CRN}{\faGithub} \\
& RCFusion \cite{zheng2023rcfusion} & T-IV 2023    & SSF  & OD & View-of-Delft/TJ4DRadSet & -- \\
& Redformer \cite{cui2023redformer} & T-IV 2023    & SSF  & OD  & NuScenes & -- \\
& CRFNet \cite{nobis2019deep}         & SDF 2019   & MSF  & OD       & NuScenes & \href{https://github.com/TUMFTM/CameraRadarFusionNet}{\faGithub} \\
& Rvnet \cite{john2019rvnet}          & PSIVT 2019 & MSF  & OD       & NuScenes & \href{https://github.com/patrick-llgc/Learning-Deep-Learning}{\faGithub} \\
& CenterFusion \cite{nabati2021centerfusion} & WACV 2021  & MSF  & OD       & NuScenes & \href{https://github.com/mrnabati/CenterFusion}{\faGithub} \\
& Radiant \cite{long2023radiant}        & AAAI 2021 & MSF  & OD       & NuScenes & \href{https://github.com/longyunf/radiant}{\faGithub} \\
& RCBEV \cite{zhou2023bridging}     & T-IV 2023    & MSF  & OD       & NuScenes & -- \\
& LXL \cite{xiong2023lxl}     &  T-IV 2023  & MSF  & OD       & View-of-Delft/TJ4DRadSet & \href{https://github.com/XiongWeiyi/LXL}{\faGithub} \\
& SparseFusion3d \cite{yu2023sparsefusion3d} & T-IV 2023  & MSF  & OD  & NuScenes & -- \\
& BEVCar \cite{schramm2024bevcar}  & IROS 2024     & MSF  & OD       & NuScenes & \href{https://github.com/robot-learning-freiburg/BEVCar}{\faGithub} \\
\midrule
\multirow{12}{*}{C\&L}
& BEVFusion \cite{liu2023bevfusion}
& NeurIPS 2022 & SSF  & OD    & NuScenes & \href{https://github.com/ADLab-AutoDrive/BEVFusion}{\faGithub} \\
& BEVFusion-MIT \cite{liang2022bevfusion}
& ICRA 2023    & SSF  & MapSeg/OD  & NuScenes & \href{https://github.com/mit-han-lab/bevfusion}{\faGithub} \\
& UVTR \cite{li2022unifying}
& NeurIPS 2023 & SSF  & OD         & NuScenes & \href{https://github.com/dvlab-research/UVTR}{\faGithub} \\
& FusionFormer \cite{hu2023fusionformer}
& arXiv 2023   & SSF  & OD         & NuScenes & \href{https://github.com/ADLab-AutoDrive/FusionFormer}{\faGithub} \\
& MV3D \cite{chen2017multi}            & CVPR 2017    & MSF  & OD         & ATG4D & \href{https://github.com/bostondiditeam/MV3D}{\faGithub} \\
& PointAugmenting \cite{wang2021pointaugmenting}
& CVPR 2021    & MSF  & OD         & NuScenes & \href{https://github.com/VISION-SJTU/PointAugmenting}{\faGithub} \\
& SDVTracker \cite{gautam2021sdvtracker}
& ICCV 2021    & MSF  & OD         & ATG4D & -- \\
& Fast-CLOCs \cite{pang2022fast}       
& WACV 2022  & MSF  & OD  & ATG4D/NuScenes & -- \\
& TransFusion \cite{bai2022transfusion}
& CVPR 2022    & MSF  & OD & NuScenes/WOD & \href{https://github.com/VISION-SJTU/PointAugmenting}{\faGithub} \\
& AutoAlign \cite{chen2022autoalign}
& arXiv 2022   & MSF  & OD & ATG4D/NuScenes & -- \\
& AutoAlignV2 \cite{chen2022deformable}      
& ECCV 2022    & MSF  & OD   & NuScenes & \href{https://github.com/zehuichen123/AutoAlignV2}{\faGithub} \\
& BEVFusion4D \cite{cai2023bevfusion4d}
& arXiv 2023   & MSF  & OD   & NuScenes &-- \\
\midrule
\multirow{2}{*}{L\&R}
& RadarNet \cite{yang2020radarnet}     & ECCV 2020 & SSF & OD & NuScenes/DenseRadar & \href{https://github.com/Toytiny/RadarNet-pytorch}{\faGithub} \\
& InterFusion \cite{wang2022interfusion}   
& IROS 2022  & MSF  & OD & Astyx HiRes & \href{https://github.com/adept-thu/InterFusion}{\faGithub} \\
\midrule
\multirow{5}{*}{\shortstack{C\&L\&R}}
& HydraFusion \cite{malawade2022hydrafusion}  
& ICCPS 2022 & SSF  & OD   & RADIATE & \href{https://github.com/aicps/hydrafusion}{\faGithub} \\
& FUTR3D \cite{chen2023futr3d}  
& CVPR 2023  & SSF  & OD   & NuScenes & \href{https://github.com/Tsinghua-MARS-Lab/futr3d}{\faGithub} \\
& BEVGuide \cite{man2023bev} 
& CVPR 2023  & SSF  & OD   & NuScenes & \href{https://github.com/YunzeMan/BEVGuide}{\faGithub} \\
& CLR-BNN \cite{ravindran2022camera}
& CVPR 2023  & SSF  & OD   & NuScenes &-- \\
& Hidden Gems \cite{ding2023hidden}   
& CVPR 2023  & MSF & MapSeg & View-of-Delft & \href{https://github.com/Toytiny/CMFlow}{\faGithub} \\
\midrule
\multirow{12}{*}{\shortstack{Temporal  Fusion}}
& FIERY \cite{hu2021fiery}  
& ICCV 2021  & SSF  & OD  & NuScenes/Lyft & \href{https://github.com/wayveai/fiery/}{\faGithub} \\
& BEVerse \cite{zhang2022beverse}
& arXiv 2022 & SSF  & MapSeg/OD  & NuScenes & \href{https://github.com/zhangyp15/BEVerse}{\faGithub} \\
& BEVENet \cite{li2023towards}
& ITSC 2023  & SSF  & OD         & NuScenes & -- \\
& BEVDet4D \cite{huang2022bevdet4d}
& arXiv 2023 & SSF  & OD         & NuScenes &\href{https://github.com/ChenControl/BEVDet4D}{\faGithub} \\
& BEVFormer \cite{li2022bevformer}
& ECCV 2022  & MSF  & MapSeg/OD  & NuScenes &\href{https://github.com/fundamentalvision/BEVFormer}{\faGithub} \\
& ST-P3 \cite{hu2022st}
& ECCV 2022  & MSF  & MOT     & NuScenes & \href{https://github.com/OpenDriveLab/ST-P3}{\faGithub} \\
& PETRv2 \cite{liu2023petrv2}
& ICCV 2022 & MSF& MapSeg  & NuScenes/OpenLane & \href{https://github.com/megvii-research/PETR}{\faGithub} \\
& DfM \cite{wang2022monocular}     
& ECCV 2022  & MSF  & OD  & KITTI & \href{https://github.com/Tai-Wang/Depth-from-Motion}{\faGithub} \\
& StreamPETR \cite{wang2023exploring}
& ICCV 2023  & MSF  & OD   & NuScenes & \href{https://github.com/exiawsh/StreamPETR}{\faGithub} \\
& SOLOFusion \cite{park2022time}
& arXiv 2023 & MSF  & OD   & NuScenes & \href{https://github.com/Divadi/SOLOFusion}{\faGithub} \\
& DORT \cite{qing2023dort}   
& arXiv 2023 & MSF  & OD/MOT& NuScenes & \href{https://github.com/OpenRobotLab/DORT}{\faGithub} \\
& Fast-BEV \cite{li2024fast}      
& TPAMI 2024 & MSF  & OD  & NuScenes & \href{https://github.com/Sense-GVT/Fast-BEV}{\faGithub} \\
\bottomrule
\end{tabular}}
\vspace{1mm}
\begin{minipage}{\linewidth}
\footnotesize
\end{minipage}
\end{table*}

 \begin{figure*}[htbp]
    \centering
    \includegraphics[width=\textwidth]{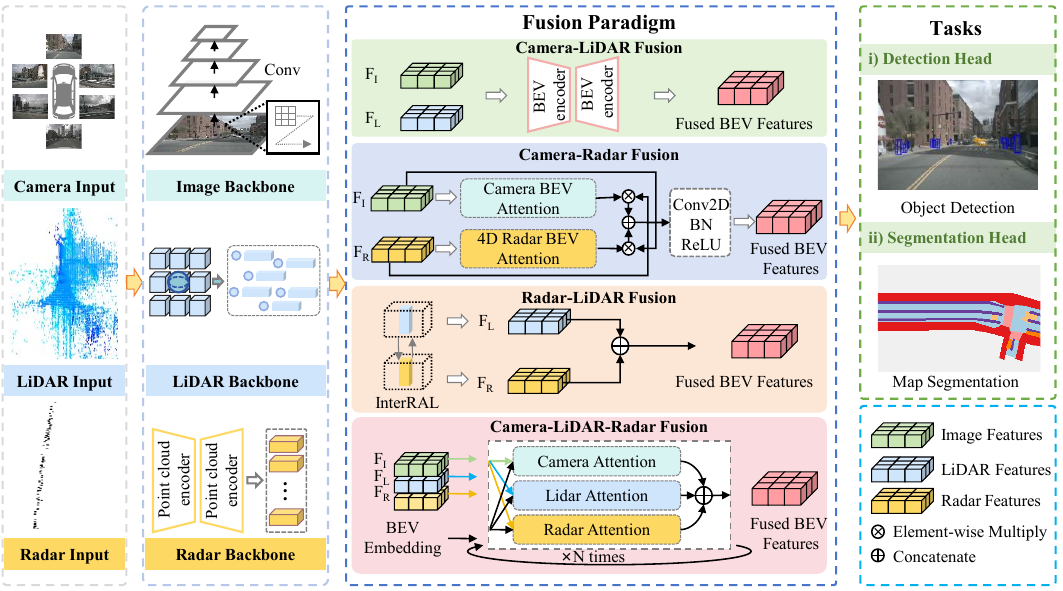}
    \caption{The overall framework of SafeBEV 2.0. SafeBEV 2.0 introduces a multi-modal vehicle-side BEV perception framework that integrates data from cameras, LiDAR, and radar. It supports four fusion paradigms: Camera–LiDAR, Camera–Radar, Radar–LiDAR, and Camera–LiDAR–Radar fusion. By leveraging the complementary advantages of heterogeneous sensors, SafeBEV 2.0 effectively mitigates the limitations of single-modality perception, significantly enhancing robustness, accuracy, and environmental understanding.}

    \label{04_BEV_Fusion}
\end{figure*}

\subsection{Camera-radar Fusion}
Cameras offer high-resolution semantics ideal for object recognition but degrade under poor visibility. Millimeter-wave radar provides robust ranging in adverse conditions but lacks semantic detail. Their fusion enhances robustness and scene understanding, as illustrated in the BEV fusion pipeline in Fig.~\ref{04_BEV_Fusion}.

\textit{\textbf{Single-stage fusion methods.}} 
CRAFT~\cite{kim2023craft} combines radar echoes with image-detected boxes to improve detection with minimal computational cost, though it underutilizes radar features. CRN~\cite{kim2023crn} projects image features into BEV and concatenates them with radar features, improving geometric alignment but relying heavily on depth estimation accuracy. RCFusion~\cite{zheng2023rcfusion} aligns FPN-derived image features with radar BEV using attention, though it may suffer from view misalignment. Redformer~\cite{cui2023redformer} applies cross-modal attention in BEV space, enhancing fusion quality at the cost of increased model complexity.

\textit{\textbf{Multi-stage fusion methods.}} CRFNet~\cite{nobis2019deep} employs early multi-scale fusion but lacks cross-modal richness. CenterFusion~\cite{nabati2021centerfusion} guides radar extraction with image boxes, effective but image-dependent. Rvnet~\cite{john2019rvnet} uses parallel sub-modules with intermediate interactions, improving spatial prediction at higher latency. Radiant~\cite{long2023radiant} refines box offsets for cross-modal consistency but adds complexity. BEVCar~\cite{schramm2024bevcar} leverages radar for image completion in BEV but suffers from transformation errors. RCBEV~\cite{zhou2023bridging} and SparseFusion3D~\cite{yu2023sparsefusion3d} enhance 3D detection via cross-view transformers and sparse completion, trading efficiency for performance. LXL~\cite{xiong2023lxl} improves depth estimation with radar but remains noise-sensitive.

\subsection{Camera-LiDAR Fusion}

Images and LiDAR provide complementary information—images offer rich textures, while LiDAR delivers precise spatial data. Their effective fusion significantly boosts perception accuracy and robustness.

\textit{\textbf{Single-stage fusion methods.}} BEVFusion~\cite{liu2023bevfusion} projects separately encoded features into a unified BEV space, streamlining the process but limiting deep feature interaction. BEVFusion-MIT~\cite{liang2022bevfusion} introduces lightweight encoders and compression to lower complexity and latency, enhancing deployment efficiency but reducing flexibility. UVTR~\cite{li2022unifying} uses a Transformer to unify BEV features, achieving strong results but showing sensitivity to data scale and tuning. FusionFormer~\cite{hu2023fusionformer} leverages a multi-scale Transformer and modality attention to improve accuracy and robustness, though its Transformer backbone increases latency and parameters. While suitable for real-time applications, single-stage methods often struggle to capture fine-grained feature interactions in complex scenes.

\textit{\textbf{Multi-stage fusion methods.}} MV3D~\cite{chen2017multi} initiates BEV fusion across scales but is constrained by early techniques. PointAugmenting~\cite{wang2021pointaugmenting} refines point features via image-guided attention, enhancing sparse-scene detection while increasing computation and reliance on images. SDVTracker~\cite{gautam2021sdvtracker} achieves deep fusion for tracking with spatiotemporal consistency, though limited by data and efficiency issues. TransFusion~\cite{bai2022transfusion} applies Transformer-based fusion for small object detection at high cost. AutoAlign~\cite{chen2022autoalign} and AutoAlignV2~\cite{chen2022deformable} learn alignments to address spatial/semantic mismatches and improve generalization. BEVFusion4D~\cite{cai2023bevfusion4d} adds spatiotemporal fusion for 4D perception but demands high resources. Fast-CLOCs~\cite{pang2022fast} uses cascaded fusion to balance speed and accuracy for real-time use.

\subsection{Radar-LiDAR Fusion}
LiDAR offers high-precision 3D geometry essential for accurate localization and mapping, but its performance degrades under adverse weather, poor lighting, and at long distances. In contrast, millimeter-wave radar maintains robustness in such conditions and provides reliable velocity measurements, making it a strong complement to LiDAR~\cite{yang2024ralibev}. Fusing the two modalities enables more resilient and comprehensive perception across diverse driving environments, as illustrated in Fig.~\ref{04_BEV_Fusion}.

\textit{\textbf{Single-stage fusion methods.}}
RadarNet~\cite{yang2020radarnet} conducts early voxel-level fusion of radar and LiDAR features to enhance long-range object detection. It further applies attention-based late fusion to improve radar–object association without relying on dense annotations. However, its performance is sensitive to voxel resolution and scene variability, and it remains susceptible to radar-induced noise.

\textit{\textbf{Multi-stage fusion methods.}} 
InterFusion~\cite{wang2022interfusion} proposes a lightweight architecture with angular compensation and cross-modal matching to align radar and LiDAR data progressively, reducing both computational cost and fusion artifacts. Nonetheless, its effectiveness depends on preprocessing quality and lacks semantic-level feature modeling, limiting robustness in complex scenarios.

\subsection{Camera-LiDAR-radar Fusion}

With the rapid advancement of autonomous driving, the fusion of multimodal sensor data from cameras, LiDAR, and radar has become essential for enhancing perception accuracy and robustness. Each modality offers distinct advantages and suffers from inherent limitations. Therefore, well-designed fusion strategies are critical for achieving reliable performance in complex and dynamic environments.

\textit{\textbf{Single-stage fusion methods.}} 
HydraFusion~\cite{malawade2022hydrafusion} adopts a dynamic fusion strategy that adaptively selects early, middle, or late fusion based on driving scenarios, improving robustness in complex conditions. However, its dependence on accurate scene understanding increases system complexity. FUTR3D~\cite{chen2023futr3d} introduces a Transformer with a modality-agnostic sampler for end-to-end 3D detection across heterogeneous sensors, offering strong flexibility but incurring high computational cost and degraded real-time performance under sparse data or adverse environments. BEVGuide~\cite{man2023bev} employs location-aware, sensor-agnostic attention for BEV-space fusion without spatial warping, enhancing efficiency. Nonetheless, its fixed BEV partitioning may impair representation quality, as suboptimal division can lead to local information loss.

\textit{\textbf{Multi-stage fusion methods.}} CLR-BNN~\cite{ravindran2022camera} leverages Bayesian neural networks to integrate camera, LiDAR, and radar features, enhancing uncertainty estimation and robustness under varying conditions, albeit with increased computational cost and compromised real-time performance. Hidden Gems~\cite{ding2023hidden} proposes a multi-stage cross-modal supervision framework, utilizing depth and flow consistency to guide radar representation learning within a radar scene flow paradigm. While it improves semantic and motion modeling—especially for distant or small objects—it introduces training complexity and requires high-quality cross-modal supervision.

\subsection{Temporal Fusion}
Temporal fusion enhances dynamic scene understanding and object detection by integrating spatial–temporal features across frames. Methods are categorized into single-stage, which offer efficiency but limited temporal depth, and multi-stage, which improve temporal modeling at higher computational cost.

\textit{\textbf{Single-stage fusion methods.}}
These methods fuse BEV features across frames in a unified step, often leveraging attention mechanisms for efficiency. MotionNet~\cite{wu2020motionnet} adopts a spatio-temporal pyramid to extract context from multi-frame point clouds. FIERY~\cite{hu2021fiery} aggregates sequential BEV features for future prediction via 3D fusion. BEVerse~\cite{zhang2022beverse} aligns historical BEV features before fusion to address ego-motion. BEVDet4D~\cite{huang2022bevdet4d} improves velocity estimation by cross-frame fusion. StreamPETR~\cite{wang2023exploring} uses a query-based approach for temporal propagation, while BEVENet~\cite{li2023towards} designs a specialized module for temporally aware representation.

\textit{\textbf{Multi-stage fusion methods.}}
These approaches aggregate temporal features across frames or scales for improved association. STA-ST~\cite{saha2021enabling} performs multi-scale fusion to enhance dynamic understanding, while ST-P3~\cite{hu2022st} models past and future states via a dual-path design. SOLOFusion~\cite{park2022time} aligns and iteratively fuses BEV features for frame consistency. Fast-BEV~\cite{li2024fast} uses a Fast-Ray transform to efficiently project features to multi-scale BEV. BEVFormer~\cite{li2022bevformer} introduces a query-based framework for flexible interaction, while PETRv2~\cite{liu2023petrv2} and UniFusion~\cite{qin2023unifusion} integrate ego- and object-motion via cross-attention. DfM~\cite{wang2022monocular} and DORT~\cite{qing2023dort} further refine representations using inter-frame matching or recurrent fusion.

\subsection{Challenges and Limitations}\label{II-D}
Multimodal fusion improves safety and robustness in autonomous driving by leveraging the complementary strengths of cameras, LiDAR, and radar, mitigating limitations such as lighting sensitivity and sparse returns. However, key challenges persist. Precise calibration and synchronization are required to align heterogeneous sensors with differing resolution and noise characteristics—misalignment degrades downstream tasks. Adverse weather, occlusion, and inconsistent visibility further impact sensor reliability and complicate fusion. Real-time inference is also constrained by computational limitations, demanding an accuracy–efficiency trade-off. Additionally, onboard fusion struggles with occluded or rare objects due to its restricted field of view. collaborative perception via V2X can alleviate this by sharing data among vehicles and infrastructure, extending perception range and improving situational awareness.

% \subsection{This is an example for second level head---subsection head}\label{subsec2}

% \subsubsection{This is an example for third level head---subsubsection head}\label{subsubsec2}

% Sample body text. Sample body text. Sample body text. Sample body text. Sample body text. Sample body text. Sample body text. Sample body text. 

\section{SafeBEV 3.0: BEV perception method based on multi-agent collaboration} \label{sec4}

\begin{figure*}[t]
    \centering
    \includegraphics[width=\textwidth]{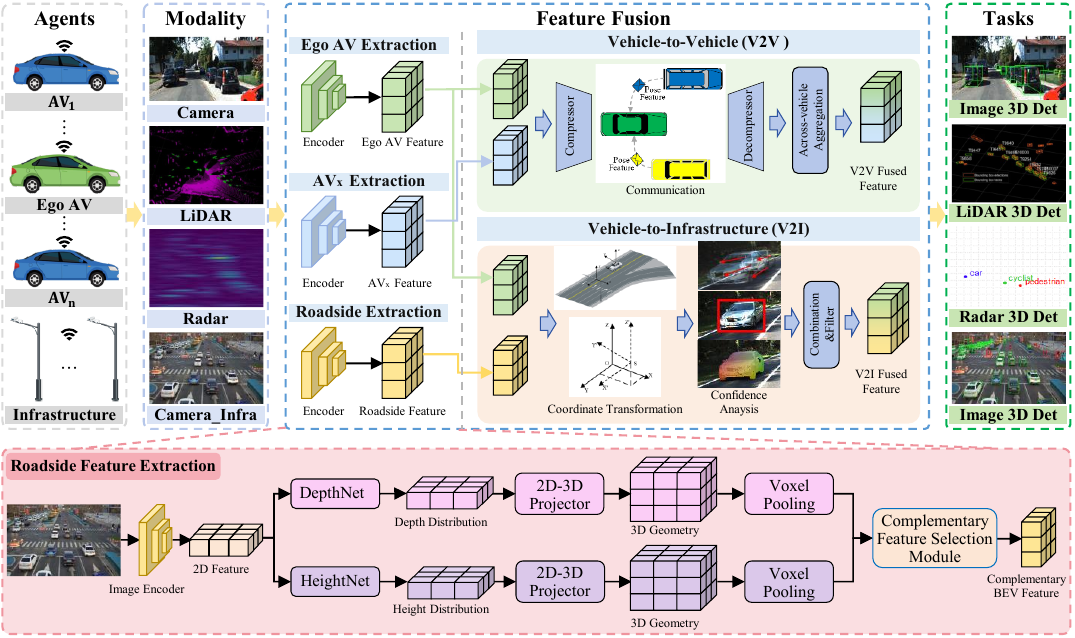}
    \caption{The overall framework of SafeBEV 3.0. The framework incorporates multi-agent collaboration among ego and partner autonomous vehicles, alongside roadside infrastructure. By fusing multimodal sensory inputs from cameras, LiDAR, and radar, the system enables more comprehensive and robust BEV perception.}
    \label{section4_framework}
\end{figure*}

With the evolution from SafeBEV 1.0 to SafeBEV 2.0, vehicle-side BEV perception has achieved improved robustness. However, it remains constrained by occlusion, limited coverage, and sensor layout. SafeBEV 3.0 advances this paradigm toward multi-agent collaboration, leveraging roadside infrastructure and inter-vehicle communication to enhance spatial coverage and system redundancy. This section is organized as follows: Section~\ref{section4A} introduces infrastructure-assisted BEV perception for occlusion mitigation; Section~\ref{section4B} reviews collaborative BEV approaches across connected agents; and Section~\ref{section4C} analyzes key challenges and system-level constraints. An overview is shown in Fig.~\ref{section4_framework}.

\subsection{Roadside BEV Perception Methods} \label{section4A}

\begin{figure}[t]
\centering
\includegraphics[scale=0.33]{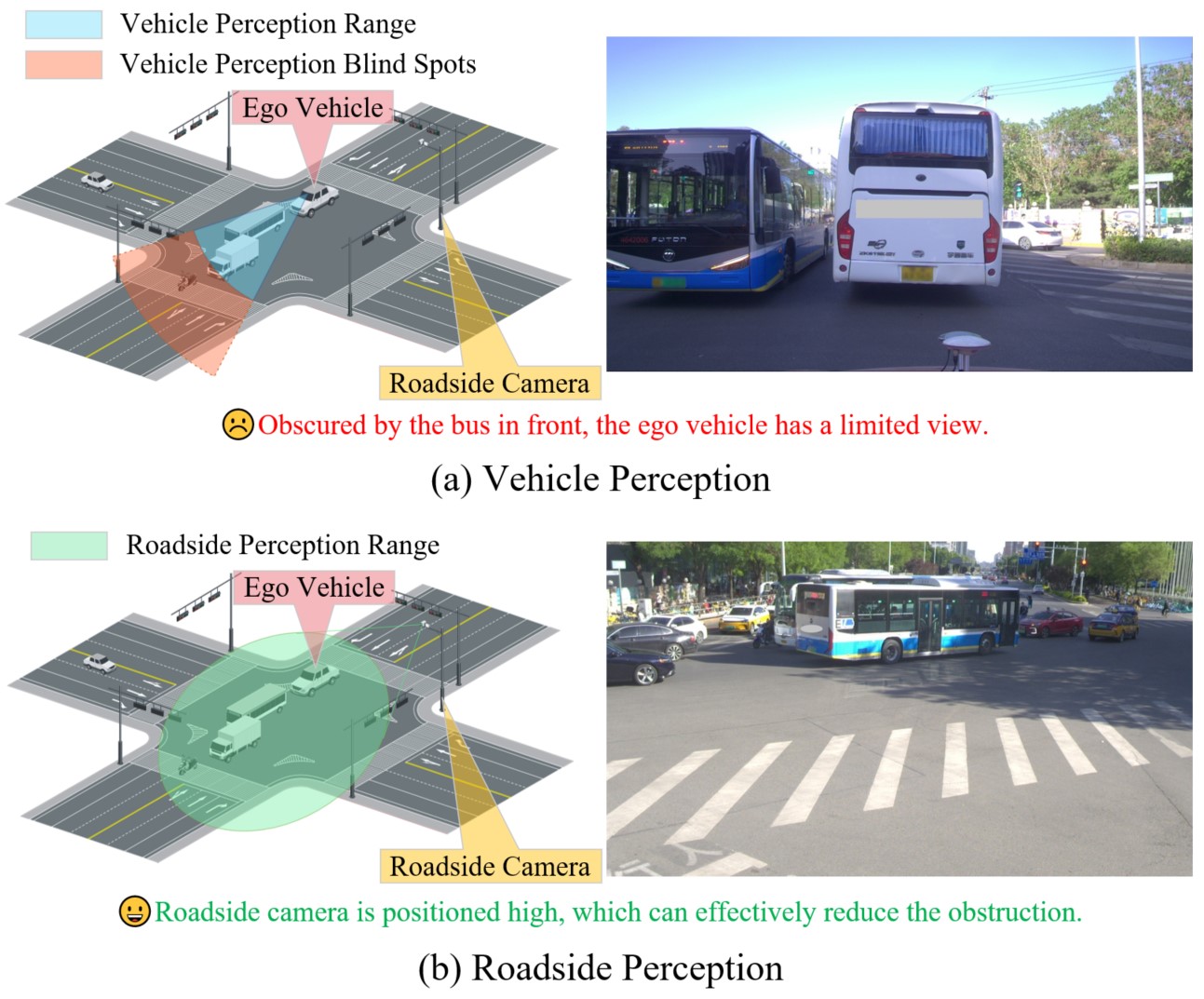}
\caption{Comparison between vehicle-side and roadside perception perspectives. In (a) vehicle perception, the ego vehicle’s sensors are obstructed by a preceding bus, resulting in a limited field of view. In contrast, (b) roadside perception illustrates how a high-mounted roadside sensor provides a broader and less occluded BEV perception, effectively mitigating the blind spot caused by occlusion.}
\label{fig6}
\end{figure}

Roadside BEV perception leverages fixed, elevated sensors such as cameras and LiDAR to detect the categories, position, velocity, and orientation of traffic participants in real time. Compared to vehicle-side sensors, roadside setups offer superior robustness against occlusion and provide broader, more stable coverage, as shown in Fig.~\ref{fig6}. 
This section introduces representative methods across three modality types—BEV Camera, BEV LiDAR, and BEV Fusion—summarized in Table~\ref{chapter4} by sensor type, task, and code availability.

\subsubsection{BEV Camera} \label{section4.1.1}

To extend perception range while ensuring robustness and accuracy, camera-based roadside BEV solutions have garnered increasing attention. These methods benefit from the maturity and cost-effectiveness of camera technology, as well as the fixed installation of roadside equipment, which minimizes dynamic calibration errors. By leveraging visual data captured from roadside infrastructure, such systems typically perform object detection and lane estimation via a pipeline comprising image encoding, 2D-to-3D projection, and voxel-based spatial reasoning.

Conventional depth-based BEV approaches often suffer from inaccurate 2D-to-3D mappings. To address this, BEVHeight \cite{yang2023bevheight} uses the height of roadside cameras to predict per-pixel elevation, thereby improving detection robustness. Furthermore, BEVHeight++ \cite{yangBEVHeightRobustVisual2023} integrates depth and height cues to further enhance geometric reliability. CoBEV \cite{shiCoBEVElevatingRoadside2024} advances toward end-to-end BEV detection, demonstrating robustness under long-range conditions, camera noise, and scene parameter variations. BEVSpread \cite{wangBEVSpreadSpreadVoxel2024} proposes an improved voxel pooling mechanism that reduces localization errors by addressing approximation biases. For multi-camera systems, CBR~\cite{fanCalibrationfreeBEVRepresentation2023} introduces a calibration-free BEV representation framework to handle installation variability and calibration noise, while RopeBEV~\cite{jiaRopeBEVMultiCameraRoadside2024} further addresses sparse perception issues in multi-view fusion.

\textit{\textbf{Accuracy–Robustness–Safety Analysis:}} Camera-based roadside perception alleviates key limitations of vehicle-side systems, such as occlusion and narrow fields of view. By expanding observable areas and reducing blind spots, it enhances both safety and perception redundancy. However, purely visual methods are susceptible to environmental degradation and lack active sensing, limiting adaptability and range accuracy. Consequently, LiDAR is often regarded as an essential complement to ensure reliability in safety-critical scenarios.

\subsubsection{BEV LiDAR} \label{section4.1.2}

LiDAR offers inherent advantages over camera-based sensors in roadside BEV perception. By capturing high-precision 3D point clouds, it provides accurate spatial geometry and maintains consistent performance under diverse lighting conditions where cameras typically struggle, thereby enhancing perceptual reliability and safety.

A key challenge in LiDAR-based roadside BEV perception is the reliable separation of dynamic objects from complex static backgrounds. To address background interference, Zhang et al.~\cite{zhangRoadsideLidarVehicle2022} utilize intensity and range information to extract static scene features. Cui et al.~\cite{cuiAutomaticVehicleTracking2019} improve lane detection by incorporating LiDAR-based structural cues, while Wu et al.~\cite{wuAutomaticBackgroundFiltering2017} extend detection range and reduce computational overhead. For low-density LiDAR data, Lin et al.~\cite{linAutomaticLaneMarking2021} introduce a ground segmentation algorithm that enhances lane marking recognition by accurately identifying ground points. Building upon these foundations, CetrRoad~\cite{shiCenterAware3DObject2023} adopts a deformable cross-attention mechanism to improve object-level perception, achieving state-of-the-art performance. Moreover, infrastructure-mounted LiDAR systems have enabled real-time analytics for multi-class targets, including pedestrian and vehicle localization, speed estimation, and directional tracking~\cite{zhaoDetectionTrackingPedestrians2019a}, offering critical support for roadside intelligence.

\textit{\textbf{Accuracy–Robustness–Safety Analysis:}} 
Although still in early deployment, LiDAR-based roadside BEV perception demonstrates superior depth estimation and long-range tracking capabilities. However, unimodal reliance on LiDAR remains vulnerable to performance degradation under adverse weather, limiting robustness. Furthermore, its high cost and deployment complexity constrain scalability for large-scale implementation.

\subsubsection{BEV Fusion} \label{section4.1.3}

Roadside BEV perception systems based on single sensing modalities face inherent limitations that compromise perception accuracy and robustness. To overcome these challenges, multimodal fusion has become a critical approach by integrating complementary information from heterogeneous sensors, enabling more reliable and comprehensive environmental understanding.

Roadside BEV fusion methods can effectively overcome the drawbacks of unimodal systems. For instance, BEVRoad~\cite{BEVRoadCrossModalTemporaryRecurrent} employs cross-modal fusion to achieve accurate velocity and localization estimates, even under challenging conditions, while also improving occlusion handling. HSRDet~\cite{chenAccurateRobustRoadside2024} enhances perception fidelity by constructing detailed scene representations and utilizing attention-based fusion to produce robust BEV features. Furthermore, a fusion-based tracking framework~\cite{wangObjectTrackingBased2022} integrates attention mechanisms to improve speed estimation, tracking range, trajectory recovery, and resilience against object loss. Collectively, these works underscore the importance of camera–LiDAR fusion for achieving reliable and comprehensive roadside BEV perception.

\textit{\textbf{Accuracy–Robustness–Safety Analysis:}} Multimodal fusion substantially strengthens the robustness and accuracy of roadside BEV perception by exploiting the complementary strengths of cameras and LiDAR. However, ensuring the safety and integrity of fusion-based systems introduces significant technical challenges. These include robust fault tolerance to sensor failures or corrupted data, and precise spatio-temporal calibration across disparate sensor data. Consequently, advancing adaptive and resilient fusion algorithms remains a critical research direction for developing safe and scalable roadside BEV perception systems.

\subsubsection{Challenges and Limitations} \label{section4.1.4}

Despite its safety potential, roadside BEV perception faces inherent challenges. Camera-based methods are sensitive to lighting and weather, and lack precise depth estimation. LiDAR-only approaches offer geometric accuracy but provide sparse data and limited semantics under adverse conditions. While multimodal fusion improves robustness by leveraging complementary sensors, it increases complexity in synchronization, calibration, and fault tolerance. Additionally, the fixed viewpoints of roadside systems constrain perception coverage, limiting situational awareness. These challenges underscore the need for collaborative BEV perception to achieve broader and more resilient environmental understanding through multi-agent collaboration.

\subsection{Collaborative BEV Perception Methods} \label{section4B}

To overcome the limitations of vehicle-side BEV systems—such as occlusions, narrow fields of view, and reduced robustness—collaborative BEV perception enables real-time information exchange among vehicles and infrastructure. This paradigm significantly expands spatial coverage, improves detection accuracy, and enhances system-level resilience.

\begin{figure}[!t]
    \centering
    \includegraphics[scale=1]{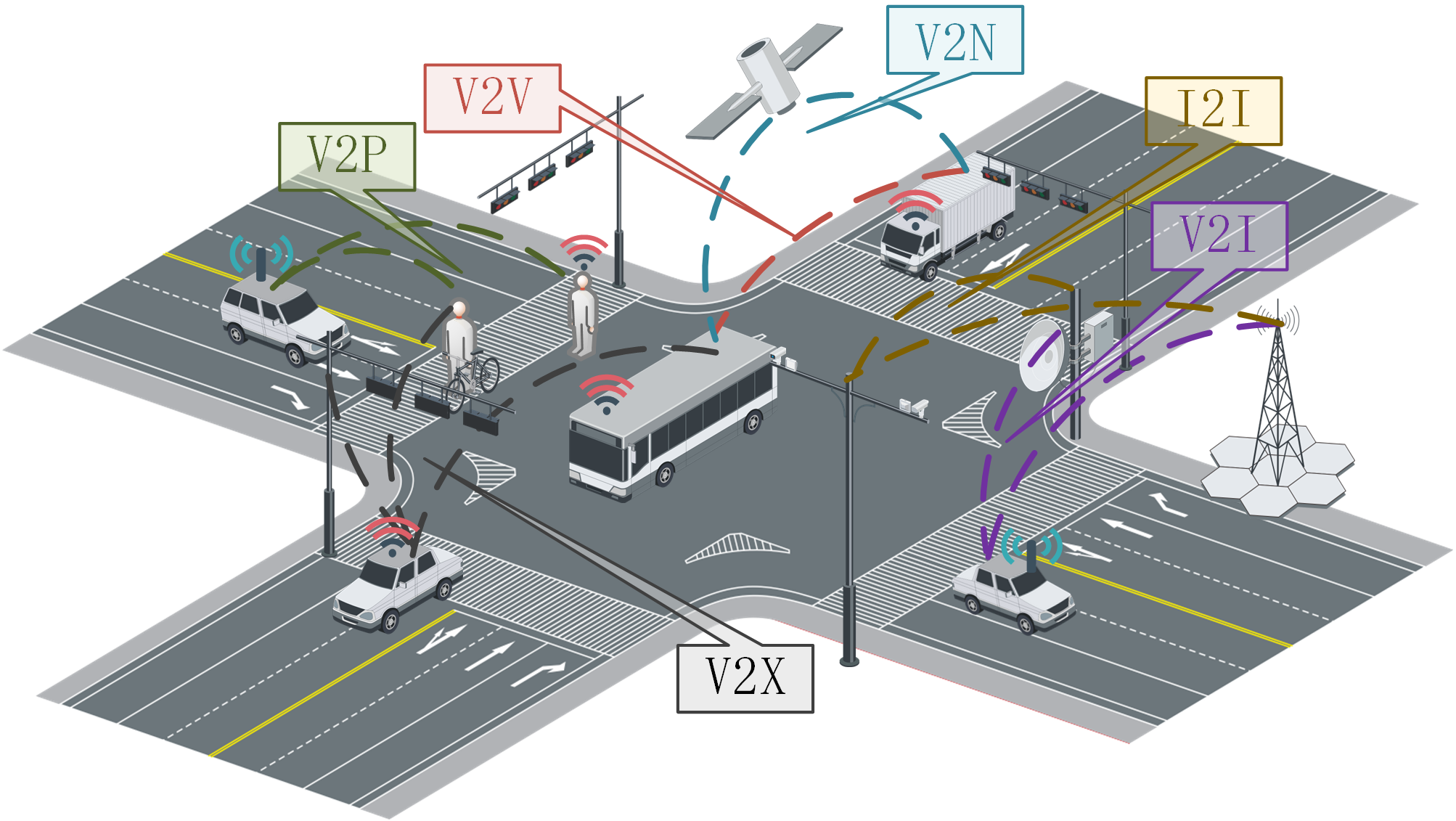}
    \caption{Overview of collaborative perception in SafeBEV 3.0. The scenario integrates V2V, V2I, V2P, V2N, and I2I communication to enable shared perception across vehicles, infrastructure, and other agents, enhancing coverage, redundancy, and situational awareness.}
    \label{fig:v2x_pic}
\end{figure}

As illustrated in Fig.~\ref{fig:v2x_pic} and summarized in Table~\ref{chapter4}, collaborative frameworks are categorized by agent type: V2V, V2I, I2I, and V2X. Each supports both single-modal and multi-modal sensor fusion across different agents—referring respectively to the fusion of homogeneous or heterogeneous data collected from multiple vehicles and infrastructure nodes. The following subsections review state-of-the-art approaches across three representative paradigms—V2V, V2I, and V2X/I2I—focusing on their contributions to accuracy, robustness, and safety.

\subsubsection{V2V Collaborative BEV Perception} \label{section4.2.1}
V2V collaborative BEV perception serves as a pivotal solution to address the fundamental limitations of ego-centric perception, notably occlusion, restricted sensing range, and insufficient environmental redundancy. By enabling real-time information exchange among multiple vehicles, V2V frameworks establish decentralized collaborative networks that significantly expand spatial perception, enhance long-range detection in non-line-of-sight scenarios, and improve the completeness and robustness of BEV representations.

\textit{Single-modal V2V Fusion:} Early work in this area has focused on fusing homogeneous sensor data, such as camera images or LiDAR point clouds, among vehicles. In vision-based systems, Transformer architectures are increasingly adopted for their global context modeling, as in CoBEVT~\cite{xuCoBEVTCooperativeBirds2022} with axial attention for efficient multi-agent BEV fusion, and TempCoBEV~\cite{rossleUnlockingInformationTemporal2024}, which incorporates temporal modeling to mitigate communication latency and misalignment. CoCa3D~\cite{hu2023collaboration} further demonstrates the benefits of collaborative depth estimation for distant object detection. LiDAR-based frameworks emphasize geometric alignment and temporal consistency; for instance, CoBEVFlow~\cite{weiAsynchronyRobustCollaborativePerception2023a} compensates for temporal asynchrony via motion modeling, while V2VNet~\cite{wangV2VNetVehicletoVehicleCommunication2020} leverages graph neural networks for structured feature aggregation. LCRN-V2VAM~\cite{liLearningVehicletoVehicleCooperative2023} extends these approaches with uncertainty-aware attention and feature repair modules to enhance robustness under degraded communication.

\textit{Multi-modal V2V Fusion:} To further enhance perception accuracy and resilience, recent work has explored multimodal V2V fusion that jointly leverages heterogeneous sensor data across vehicles. These approaches combine the semantic richness of visual inputs with the geometric precision of LiDAR, resulting in more comprehensive BEV scene understanding. V2VFormer++~\cite{yinV2VFormerMultiModalVehicletoVehicle2024} proposes a global-local Transformer framework with dynamic channel fusion to enable scalable and effective multimodal aggregation. CoBEVFusion~\cite{qiaoCoBEVFusionCooperativePerception2023} adopts a dual-window cross-attention mechanism to simultaneously boost semantic segmentation and 3D detection. HM-ViT~\cite{xiangHMViTHeteromodalVehicletoVehicle2023} introduces a heterogeneous-modality Transformer designed for flexible sensor alignment, while MCoT~\cite{shiMCoTMultiModalVehicletoVehicle2023} exploits geometric priors to refine BEV feature alignment and fusion granularity.

\textit{Accuracy–Robustness–Safety Analysis:} V2V collaborative BEV perception significantly advances spatial completeness, detection accuracy, and system robustness by harnessing inter-vehicle collaboration. Single-modal approaches offer architectural simplicity and lower communication overhead but remain limited by sensor-specific constraints. In contrast, multimodal fusion provides richer representations at the cost of increased complexity in synchronization and calibration. Future research should focus on integrating V2V with broader V2X frameworks, developing latency-aware and adaptive fusion algorithms, and addressing robustness under real-world deployment conditions.

\begin{table}[!t]
\centering
% \small
\setlength{\tabcolsep}{4.3pt} 

\caption{Summary of representative methods for collaborative BEV perception across different collaboration paradigms. “View” denotes the collaboration type: Roadside(Road), Vehicle-to-Vehicle (V2V), Vehicle-to-Infrastructure (V2I), and Vehicle-to-Everything (V2X). “Type” refers to the sensing modality: “C” for camera, “L” for LiDAR, and “C\&L” for multimodal fusion. “Tasks” includes object detection (OD) and object tracking(OT). Public code availability is indicated in the last column.}

\label{chapter4}
\tabcolsep=1.6mm

\begin{tabular}{m{0.3cm}<{\centering}@{\hskip 10mm}m{2.5cm}<{\raggedright}@{\hskip 0.9mm}m{2.1cm}<{\centering}@{\hskip 1.2mm}m{0.68cm}<{\centering}m{0.8cm}<{\centering}m{0.5cm}<{\centering}<{\centering}m{3.2cm}<{\centering}}
\toprule
View & \makebox[2.1cm][c]{Method} & Venue & Type & Task & Code &Remark \\
\midrule
\multirow{11}{*}{Road}
& BEVHeight \cite{yang2023bevheight} & CVPR 2023 &C &OD & \href{https://github.com/ADLab-AutoDrive/BEVHeight}{\faGithub} &Height first used\\
& BEVHeight++ \cite{yangBEVHeightRobustVisual2023} & TPAMI 2025 &C & OD & \href{https://github.com/yanglei18/BEVHeight_Plus}{\faGithub} &Height\&depth fusion\\
& CoBEV \cite{shiCoBEVElevatingRoadside2024} & TIP 2024 &C & OD & \href{https://github.com/MasterHow/CoBEV}{\faGithub} &End-to-end fusion\\
& BEVSpread \cite{wangBEVSpreadSpreadVoxel2024} & CVPR 2024 &C & OD & \href{https://github.com/DaTongjie/BEVSpread}{\faGithub} &Spread Voxel Pooling\\
& CBR \cite{fanCalibrationfreeBEVRepresentation2023} & IROS 2023 &C & OD & \href{https://github.com/leofansq/CBR}{\faGithub} &Calibration-free\\
& RopeBEV \cite{jiaRopeBEVMultiCameraRoadside2024} & arXiv 2024 &C & OD & -- &For multi-camera\\
& Zhao et al. \cite{zhaoDetectionTrackingPedestrians2019a} & TR-C 2019 &L & OD\&OT & -- &Clustering Methods\\
& Cui et al. \cite{cuiAutomaticVehicleTracking2019} & IS 2019 &L & OD\&OT & -- &Infrastructure sense\&broadcast\\
& BEVRoad \cite{BEVRoadCrossModalTemporaryRecurrent} & EasyChair 2024 &C\&L & OD & -- &Adaptive fusion\\
& HSRDet \cite{chenAccurateRobustRoadside2024} & JSEN 2024 &C\&L & OD & --  &Height based fusion\\
& Wang et al. \cite{wangObjectTrackingBased2022} & TIM 2022 &C\&L & OD\&OT & -- &Adaptive weight\\
\midrule
\multirow{10}{*}{V2V}
& CoBEVT \cite{xuCoBEVTCooperativeBirds2022}         & CoRL 2022   & C     & OD & \href{https://github.com/DerrickXuNu/CoBEVT}{\faGithub} &Sparse Transformers\\
& TempCoBEV \cite{rossleUnlockingInformationTemporal2024} & IV 2024     & C     & OD & \href{https://github.com/cvims/TempCoBEV}{\faGithub} &Temporal embeddings\\
& CoCa3D \cite{hu2023collaboration}       & CVPR 2023   & C     & OD & \href{https://github.com/MediaBrain-SJTU/CoCa3D}{\faGithub} &Depth fusion\\
& CoBEVFlow \cite{weiAsynchronyRobustCollaborativePerception2023a} & NeurIPS 2023 & L     & OD & \href{https://github.com/MediaBrain-SJTU/CoBEVFlow}{\faGithub} &Asynchrony BEV flow\\
& V2VNet \cite{wangV2VNetVehicletoVehicleCommunication2020}        & ECCV 2020   & L     & OD & -- &--\\
& LCRN-V2VAM \cite{liLearningVehicletoVehicleCooperative2023}     & TIV 2023    & L     & OD & \href{https://github.com/jinlong17/V2VLC}{\faGithub} &Lossy communication\\
& CoBEVFusion \cite{qiaoCoBEVFusionCooperativePerception2023}     & DICTA 2024  & C\&L  & OD & -- &Dual cross-Attention\\
& HM-ViT \cite{xiangHMViTHeteromodalVehicletoVehicle2023}         & ICCV 2023   & C\&L  & OD & \href{https://github.com/XHwind/HM-ViT}{\faGithub} &Hetero-modal\\
& MCoT \cite{shiMCoTMultiModalVehicletoVehicle2023}              & ICPADS 2023 & C\&L  & OD & -- &Cross-attention\\
& V2VFormer++ \cite{yinV2VFormerMultiModalVehicletoVehicle2024}   & TITS 2023   & C\&L  & OD & -- &Global-Local  transformer\\
\midrule
\multirow{11}{*}{V2I}
& VIMI \cite{wangVIMIVehicleInfrastructureMultiview2023}          & arXiv 2023  & C     & OD & -- &Multi-view\\
& BEVSync \cite{wang2025bevsync}                                  & AAAI 2025   & C     & OD & -- &Uncertain delays\\
& VI-BEV \cite{mengVIBEVVehicleInfrastructureCollaborative2025b}  & OJITS 2025  & C     & OD & -- &Cross-attention\\
& CoFormerNet \cite{liCoFormerNetTransformerBasedFusion2024}      & Sensors 2024& L     & OD & -- &Temporal aggregation\\
& V2IViewer \cite{yi2024v2iviewer}                                & TNSE 2024   & L     & OD\&OT & -- &Efficient collaboration\\
& CenterCoop \cite{zhouCenterCoopCenterBasedFeature2024a}         & RAL 2023    & L     & OD & -- &Feature aggregation\\
& V2I-BEVF \cite{xiangV2IBEVFMultimodalFusion2023a}               & ITSC 2023   & C\&L  & OD & -- &Multi-modal fusion\\
& V2I-Coop \cite{zhou2024v2i}                                     & TMC 2024    & C\&L  & OD & -- &Accident black spots\\
& MSMDFusion \cite{jiaoMSMDFusionFusingLiDAR2023}                 & CVPR 2023   & C\&L  & OD & \href{https://github.com/sxjyjay/msmdfusion}{\faGithub} &Multi scale\&depth\\
& CO\textsuperscript{3} \cite{chenCO^3CooperativeUnsupervised2022} & ICLR 2023   & C\&L  & OD & \href{https://github.com/Runjian-Chen/CO3}{\faGithub} &Unsupervised learning\\
& VICOD \cite{yuMultistageFusionApproach2022}                     & WCMEIM 2022 & C\&L  & OD & -- &Feature Flow Net\\
\midrule
\multirow{2}{*}{V2X}
& V2X-BGN \cite{zhang2024v2x}  & IV 2024   & C     & OD & -- &Non-Maximum suppression\\
& BEV-V2X \cite{chang2023bev} & TIV 2023  & L     & OD & -- &BEV Fusion\&occupancy\\
\bottomrule
\end{tabular}
\end{table}

\subsubsection{V2I Collaborative BEV Perception}\label{section4.2.2}

V2I collaborative BEV perception leverages the complementary advantages of roadside infrastructure to enhance onboard perception systems. Unlike dynamic vehicle-mounted sensors, infrastructure devices such as elevated cameras and LiDARs offer a fixed and stable field of view, enabling persistent monitoring of blind zones and complex intersections. Through real-time information exchange with vehicles, V2I collaboration significantly improves BEV scene understanding, particularly in occlusion-heavy or cluttered urban environments where ego-perception is insufficient.

\textit{Single-modal V2I Fusion:} V2I single-modal methods align and fuse homogeneous sensor data between vehicles and infrastructure nodes to build unified BEV representations. In camera-based fusion, BEV features extracted from infrastructure and onboard images are aggregated to mitigate view occlusion and extend the visual field. VIMI~\cite{wangVIMIVehicleInfrastructureMultiview2023} introduces dynamic enhancement modules to alleviate projection loss and calibration drift, while BEVSync~\cite{wang2025bevsync} compensates for temporal desynchronization through a dedicated extractor-compensator mechanism. VI-BEV~\cite{mengVIBEVVehicleInfrastructureCollaborative2025b} further strengthens spatial feature consistency via cross-sensor interaction modeling. For LiDAR-based pipelines, CoFormerNet~\cite{liCoFormerNetTransformerBasedFusion2024} adopts spatiotemporal modulation attention to handle data latency and misalignment, whereas CenterCoop~\cite{zhouCenterCoopCenterBasedFeature2024a} achieves efficient centralized encoding to reduce communication bandwidth. V2IViewer~\cite{yi2024v2iviewer} integrates detection, compression, and alignment modules into an end-to-end architecture.

\textit{Multi-modal V2I Fusion:} To address the limitations of single-sensor systems, multimodal V2I fusion incorporates heterogeneous data sources—such as infrastructure LiDAR and vehicle-mounted cameras—to enrich BEV representations. These cross-modal pipelines combine the semantic richness of vision with the geometric precision of point clouds. V2I-BEVF~\cite{xiangV2IBEVFMultimodalFusion2023a} aligns BEV features through a dual-branch Transformer with deformable attention. MSMDFusion~\cite{jiaoMSMDFusionFusingLiDAR2023} introduces multi-depth projection and gated convolution to support multi-scale semantic integration. V2I-Coop~\cite{zhou2024v2i} focuses on risk-aware detection through fine-grained multimodal collaboration, while CO\textsuperscript{3}~\cite{chenCO^3CooperativeUnsupervised2022} leverages contrastive learning to enable unsupervised cross-modal representation alignment. VICOD~\cite{yuMultistageFusionApproach2022} implements multi-stage fusion with bounding box alignment to enhance late-stage detection robustness.

\textit{Accuracy–Robustness–Safety Analysis: }V2I collaborative perception complements vehicle sensing by offering elevated, stable viewpoints and enabling cross-agent redundancy. Compared with V2V paradigms, it provides more consistent coverage in structured environments and alleviates occlusion-related failures. Nonetheless, practical challenges persist, including limited infrastructure deployment, sensor calibration requirements, and communication latency. Future research may explore scalable V2X-hybrid frameworks that dynamically balance V2V and V2I contributions, self-calibration mechanisms under asynchronous inputs, and learning-based protocols for resilient multimodal alignment in dynamic urban environments.

\subsubsection{V2X and I2I Collaborative BEV Perception} \label{section4.2.3}

To achieve comprehensive scene understanding in dynamic traffic environments, V2X and I2I collaborative BEV perception extend the information-sharing scope beyond individual agents by incorporating both vehicles and infrastructure nodes. This paradigm enhances spatial coverage, mitigates occlusions, and provides richer contextual awareness in complex scenarios, as illustrated in Fig.~\ref{fig:v2x_pic}.

Recent frameworks have begun to explore this direction. BEV-V2X~\cite{chang2023bev} aggregates BEV representations from multiple vehicles through cloud or roadside units, enabling more accurate and globally consistent occupancy predictions. V2X-BGN~\cite{zhang2024v2x} introduces a global non-maximum suppression mechanism coupled with post-fusion refinement to improve detection performance under occluded conditions. In parallel, the H-V2X~\cite{liu2024h} and InScope~\cite{zhang2024inscope} datasets provide large-scale benchmarks tailored for highway and infrastructure-centric collaborative perception, helping address scalability and evaluation challenges in real-world settings.

\textit{Accuracy–Robustness–Safety Analysis: }While V2X and I2I collaborative BEV perception substantially improve spatial coverage and environmental understanding, they remain limited by persistent occlusions, inter-agent calibration errors, and heterogeneous sensing quality across platforms. Communication constraints, including bandwidth limitations and synchronization delays, further undermine real-time performance and fusion reliability. The static nature of infrastructure nodes may also introduce blind spots in dynamic scenes. Moreover, inconsistent detection priors among agents can impair global consistency and decision safety. These issues collectively pose significant barriers to achieving accurate, robust, and safe BEV perception under complex traffic conditions.

\subsection{Challenges and Limitations} \label{section4C}

Despite its potential, collaborative BEV perception faces key challenges in communication reliability, spatiotemporal alignment, and system scalability. V2V must manage dynamic topologies and heterogeneous sensors, while V2I suffers from limited infrastructure coverage and high deployment cost. Multimodal fusion further adds complexity in calibration and fault tolerance. Future research should focus on adaptive fusion architectures and lightweight, task-aware coordination strategies to handle partial observability, asynchronous data, and diverse traffic conditions. A unified design of perception, communication, and computation is essential to realize robust and scalable V2X collaborative perception.

\section{ Datasets for BEV Perception} \label{sec5}
In autonomous driving systems, perception performance is directly related to the overall safety and stability of the system, while high-quality datasets are essential for the development and evaluation of perception algorithms~\cite{wang2025collaborative},~\cite{liu2024survey},~\cite{gong2023sifdrivenet}. 
Therefore, well-designed datasets are particularly important for enhancing the robustness and safety of BEV perception. In this section, Section~\ref{single-vehicle datasets} introduces single-vehicle perception datasets; Section~\ref{Multi-Agent datasets} presents multi-agent collaborative perception datasets; Section~\ref{Security and Robustness Analysis} evaluates the extent to which these datasets support robustness and safety; and Section~\ref{Metrics and Benchmark} summarizes the safety-related evaluation metrics and benchmarks proposed in existing datasets.

{\LARGE
\begin{table*}[ht]
\centering
% \caption{Summary of Vehicle-Side Datasets. Modalities include Single-Modality (SM) and Multimodality (MM). Supported tasks include Semantic Segmentation (SS), Object Detection (OD), Instance Segmentation (IS), Multi-Object Tracking (OT), Depth Estimation (DE), Visual Odometry (VO), Sensor Fusion (SF), Trajectory Prediction (TP), Optical Flow (OF), Anomaly Segmentation (AS), and 3D Lane Detection (LD).}
\caption{Summary of vehicle-side perception datasets categorized by sensing modality: Single-Modality (SM) and Multimodality (MM). Supported tasks include Semantic Segmentation (SS), Object Detection (OD), Instance Segmentation (IS), Multi-Object Tracking (OT), Depth Estimation (DE), Visual Odometry (VO), Sensor Fusion (SF), Trajectory Prediction (TP), Optical Flow (OF), Anomaly Segmentation (AS), and 3D Lane Detection (LD). The table also reports dataset size, scene diversity, category diversity, and collection locations.}

\label{Vehicle_Perception_Datasets}
\begin{adjustbox}{width=\textwidth}
\begin{tabular}{@{}clllllllll@{}} 
\toprule
\multirow{2}{*}{Modalities} & \multirow{2}{*}{Datasets} & \multirow{2}{*}{PUB} & \multirow{2}{*}{Tasks} & \multicolumn{2}{c}{\textbf{Size}} & \multicolumn{2}{c}{\textbf{Diversity}} & \multirow{2}{*}{Location} & \multirow{2}{*}{Link} \\
\cmidrule(lr){5-6} \cmidrule(lr){7-8}
& & & & Frames & Annotation & Scenes & Category & & \\
\midrule
\multirow{10}{*}{SM} & CamVid\cite{CamVid}                    & Elsevier 2008         & SS,OD                  & 8K     & 15         & -         & 15       & Cambridge                     & \href{http://mi.eng.cam.ac.uk/research/projects/VideoRec/}{\faLink}          \\
& VPGNet\cite{vpgnet}                    & CVPR 2017             & SS,OD             & 20K    & 17         & 4         & 2        & Seoul                         & \href{https://github.com/SeokjuLee/VPGNet}{\faGithub}                        \\
& Lane Det\cite{Lane-Det}               & AAAI 2018             & LD,SS                  & 133K   & -          & 8         & 4        & China                         & \href{https://ojs.aaai.org/index.php/AAAI/article/view/12301}{\faLink}       \\
& Foggy-C\cite{foggy-cityscape}        & IJCV 2018             & SS,OD                  & 20K    & 550        & 5         & 19       & GTAV+MODs                     & \href{https://link.springer.com/}{\faLink}                                   \\& KMOTS\cite{Mots}                & CVPR 2019             & IS,OD,OT               & 10,870 & 65,213     & 25        & 2        & Karlsruhe               & \href{https://www.vision.rwth-aachen.de/page/mots }{\faLink} \\
 & IDD\cite{IDD}                       & CVPR 2019            & OD,SS,IS               & 10K    & 34         & -         & -        & Indian                            & \href{https://idd.insaan.iiit.ac.in/}{\faLink}  \\
 & Det\cite{Det}                       & CVPR 2019             & SS,IS                  & 5,424  & -          & -         & 2        & Wuhan                       & \href{https://spritea.github.io/DET/}{\faLink} \\
 & ACDC\cite{ACDC}                       & ICCV 2021             & SS                     & 4K     & -          & -         & 19       & Karlsruhe                     & \href{https://acdc.vision.ee.ethz.ch/ }{\faLink}                              \\
& StreetHazards\cite{Caos}                       & NeurIPS 2022          & AS              & 7K     & 13          & 6         & 250       & CARLA-V                       & \href{https://github.com/hendrycks/anomaly-seg }{\faGithub}                  \\
& BDD-Anomaly\cite{Caos}                       & NeurIPS 2022          & AS              & 8K     & 18          & 10         & 3       & USA                       & \href{https://github.com/hendrycks/anomaly-seg }{\faGithub}                  \\
& LiSV-3DLane\cite{zhao2024advancements} & ICRA 2024 & LD,SS & 20025 & - & - & - & Sydney & \href{https://github.com/RunkaiZhao/LiLaDet}{\faGithub} \\
& SynFog\cite{synfog}                   & CVPR 2024             & SS,OD     & 1,350  & 500        & 500       & 3        & CARLA-V                       & \href{-}{\faGithub}  \\
\midrule
\multirow{31}{*}{MM} &  KITTI\cite{kitti}                     & CVPR 2012             & OD,OT,OF,VO            & 15K    & 200K       & -         & 8        & Karlsruhe                & \href{http://www.cvlibs.net/datasets/kitti/}{\faLink} \\ & Cityscapes\cite{cityscapes}           & CVPR 2016             & SS,IS,OD               & 25K    & 943M       & 50        & 30       & European                 & \href{https://www.cityscapes-dataset.com/ }{\faLink} \\ & SYNTHIA\cite{synthia}                 & CVPR 2016             & SS,OD,OT               & 213K   & 13         & -         & 13       & Unity-V                  & \href{ https://synthia-dataset.net/}{\faLink} \\ & Maddern et al.\cite{oxford-robotcar} & SAGE 2017                   & SS,OD,OT,TP         & 20M    & -          & -         & -        & Oxford                   & \href{https://robotcar-dataset.robots.ox.ac.uk/ }{\faLink}\\ & KAIST\cite{KAIST} & ICRA 2018 & OD,DE & - & -  & 3 & 3 & South Korea & \href{http://multispectral.kaist.ac.kr/}{\faLink} \\ & Huang et al. \cite{apolloscape}       & CVPR 2018             & OD,SS,OT,DE,LD      & 140K   & 53         & -         & -        & Beijing                  & \href{ http://apolloscape.auto/}{\faLink} \\& A*3D\cite{A*3d}                        & arXiv 2019            & OD                     & 39K    & 230K       & -         & 7        & Singapore                & \href{https://github.com/I2RDL2/ASTAR-3D}{\faGithub} \\ & Argoverse\cite{Argoverse}             & CVPR 2019             & OD,OT               & 46K    & 993K       & 366       & 9        & USA                      & \href{https://www.argoverse.org/}{\faLink} \\& Astyx\cite{Astyx}                     & EuRAD 2019            & OD,SF                  & 500    & 3K         & -         & 7        & Germany                  & \href{ https://www.cruisemunich.de/ }{\faLink} \\& SKITTI\cite{Semantickitti}           & ICCV 2019             & SS            & 43,552 & 28         & 22        & -        & Karlsruhe                & \href{https://www.semantic-kitti.org/ }{\faGithub} \\& H3D\cite{H3D}                         & ICRA 2019             & OD,OT                  & 27K    & 1.1M       & 160       & 8        & USA                      & \href{https://usa.honda-ri.com/h3d}{\faLink} \\ & nuScenes\cite{nuscenes}              & CVPR 2019             & OD,OT                  & 40K    & 1.4M       & 1000      & 23       & USA                      & \href{https://github.com/nutonomy/nuscenes-devkit}{\faGithub} \\ & Yu et al.\cite{Bdd100k}               & CVPR 2020             & OD,SS,TP               & 100K   & 10         & -         & 48       & USA                      & \href{https://bdd-data.berkeley.edu}{\faLink} \\& VKITTI 2\cite{Virtualkitti2}          & CVPR 2020             & OT,SS               & 5      & -          & -         & -        & Unity 3D-V               & \href{https://europe.naverlabs.com/}{\faLink} \\& Cirrus\cite{Cirrus}                   & arXiv 2020            & OD                 & 6285   & -          & -         & 8        & USA                      & \href{https://developer.volvocars.com/open-datasets}{\faLink} \\& Barnes et al.\cite{oxford-radar-robotcar} & arXiv 2020        & SS,OD,OT            & 2.4M   & -          & 32        & -        & Oxford                   & \href{https://ori.ox.ac.uk/datasets/radar-robotcar-dataset }{\faLink} \\& STF\cite{STF}                         & CVPR 2020             & OD,SS,DE               & 1.4M   & 100K       & 18        & -        & Germany                  & \href{https://www.nuscenes.org/fog}{\faLink} \\ & Liao et al.\cite{Kitti-360}           & arXiv 2020            & OD,SS,IS               & 300K   & 150K       & -         & 37       & Karlsruhe                & \href{http://www.cvlibs.net/datasets/kitti-360 }{\faLink} \\ & PandaSet\cite{Pandase}                & arXiv 2020            & OD,SS                  & 8K     & -          & -         & 65       & USA                      & \href{https://pandaset.org/ }{\faLink} \\ & A2D2\cite{A2D2}                       & arXiv 2020            & OD,SS                  & 433K   & 53K        & -         & 38       & Germany                  & \href{https://www.a2d2.audi/a2d2/en.html}{\faLink} \\ & Waymo\cite{Waymoopendataset}         & CVPR 2020             & OD,OT               & 230K   & 12M        & -         & -        & USA                      & \href{http://www.waymo.com/open}{\faLink} \\ & NUPLAN\cite{nuplan}                   & arXiv 2021            & OD,OT              & 1500h  & -          & -         & 3        & USA                      & \href{https://www.nuplan.org/}{\faLink} \\
 & RADIATE\cite{radiate}                 & ICRA 2021             & OD,SF              & 44K    & 200K       & 7         & 8        & UK                       & \href{http://pro.hw.ac.uk/radiate/ }{\faLink} \\ & ONCE\cite{once}                       & arXiv 2021            & OD                     & 1M     & 417K       & -         & 5        & Beijing                  & \href{http://www.once-for-auto-driving.com.}{\faLink} \\ & AIODrive\cite{aiodrive}               & NeurIPS 2021          & OD,OT,SS,DE,TP    & -      & -          & -         & -        & CARLA-V                  & \href{http://www.aiodrive.org/}{\faLink} \\& SHIFT\cite{shift}                     & CVPR 2022             & SS,IS,OD,VO        & 250m   & -          & -         & 13       & CARLA-V                  & \href{http://www.vis.xyz/shift}{\faLink} \\ & OpenLane\cite{OPenlane}               & ECCV 2022             & LD,SU                  & 200K   & 880K       & 1000      & 14       & USA                      & \href{https://github.com/OpenPerceptionX/OpenLane }{\faGithub} \\ & ONCE-3DL\cite{Once-3d}                & CVPR 2022             & LD,LS,DE               & 211K   & -          & 16K       & -        & China                    & \href{https://once-3dlanes.github.io/}{\faLink} \\  & TJ4DRadSet\cite{zheng2022tj4dradset}               & ITSC 2022             & OD,OT                  & 7757   & -      & -      & 44      & China                      & \href{https://github.com/TJRadarLab/TJ4DRadSet}{\faGithub} \\ & Lyft L5\cite{Lyft}                    & arXiv 2023            & OD,OT,TP            & 425K   & 71K        & -         & 9        & USA                      & \href{https://github.com/RomainLITUD/Car-Following-Dataset-HV-vs-AV}{\faGithub} \\ & ZOD\cite{ZOD}                         & ICCV 2023             & OD,IS,SS               & 100K   & 446K       & -         & 171      & European                 & \href{https://zod.zenseact.com/}{\faLink} \\ & Argoverse2\cite{wilson2023argoverse}  & arXiv 2023            & OD,OT,SS           & 9M     & 14M        & -         & 40       & USA                      & \href{https://github.com/argoverse/argoverse-api}{\faGithub}  \\
& MANTruckScenes\cite{MANTruckScenes} & NeurIPS 2024 & OD,OT & 747 & 30K & 34 & 27 & Germany & \href{https://proceedings.neurips.cc/paper_files/paper/2024}{\faLink} \\
& OmniHD-Scenes\cite{zheng2024omnihd} & arXiv 2024 & OD,OT,SS & 450K & 514K & 1501 & 11 & China & \href{https://www.2077ai.com/OmniHD-Scenes}{\faLink} \\
\bottomrule
\end{tabular}
\end{adjustbox}

\vspace{1mm}
\begin{minipage}{\linewidth}
\end{minipage}
\label{your-table-label}
\end{table*}
}
\subsection{Vehicle-side BEV Datasets} \label{single-vehicle datasets}

This section reviews vehicle-side datasets for autonomous driving, categorized into single-modality and multimodality. We summarize their core characteristics, application contexts, research value, and inherent challenges, thereby establishing a basis for future development. Table~\ref{Vehicle_Perception_Datasets} presents a systematic comparison of these datasets across dimensions such as task types, collection scenarios, frame rates, and geographical coverage.

\subsubsection{Single-Modality BEV Datasets}

Single-modality vehicle-side datasets consist of sequential perception data collected from a single sensor type (e.g., camera or LiDAR) mounted on an individual vehicle. These datasets serve as fundamental tools for task-specific research in semantic segmentation and object detection, while providing a controlled platform to evaluate individual sensor performance limits. Their standardized structure enables comprehensive analysis of model behaviors, robustness bottlenecks, and failure patterns across various operational scenarios.

Several benchmark datasets address critical safety scenarios: SynFog~\cite{synfog} generates realistic fog conditions through physical scattering models; VPGNet~\cite{vpgnet} enhances lane detection under challenging weather and lighting conditions; IDD~\cite{IDD} captures complex traffic dynamics in unstructured environments; StreetHazards~\cite{Caos} evaluates generalization capability using synthetic hazards; and CamVid~\cite{CamVid} provides detailed annotations for intersection analysis. These resources enable targeted investigation of sensor-specific capabilities under diverse operational conditions.

Current single-modality datasets face significant challenges in safety validation, particularly regarding extreme condition representation. The scarcity of real-world edge cases (e.g., snowstorms, collision scenarios) and the reality gap in synthetic data limit comprehensive robustness evaluation. Future development should prioritize: (1) enhanced physics-based simulation fidelity, (2) systematic collection of rare safety-critical scenarios, and (3) integration of quantifiable safety metrics to establish rigorous validation frameworks for autonomous perception systems.

\subsubsection{Multimodality BEV Datasets}
Multimodal datasets integrate time-synchronized data from heterogeneous sensors, forming the basis for robust autonomous driving perception. Their core technological value lies in enabling sensor fusion strategies that leverage complementary modalities to overcome individual limitations—such as camera failures under low light or LiDAR degradation in fog—thereby improving system resilience across diverse environments~\cite{zhang2022openmpd}.

The evolution of datasets has marked significant milestones in autonomous driving research. KITTI~\cite{kitti} established foundational benchmarks with stereo cameras and LiDAR on structured roads. nuScenes~\cite{nuscenes} increased sensor diversity by introducing radar, 6D pose annotations, and challenging weather conditions. RADIATE~\cite{radiate} focused on low-visibility scenarios through radar-camera fusion in snow, rain, and fog. Argoverse2~\cite{Argoverse} emphasized complex urban interactions with 3D lane topology and detailed behavior annotations. Waymo~\cite{Waymoopendataset} advanced safety validation by providing large-scale urban and highway data, including rare accident cases.

Despite advances in sensor diversity and scenario coverage, key gaps remain. Ultra-adverse conditions—like whiteout snowstorms and dense fog—and critical events such as intersection crashes and emergency maneuvers are underrepresented, limiting boundary validation. Technical hurdles like inconsistent sensor calibration and privacy-restricted accident data also hinder progress. Future work should focus on: 1) Curating edge cases (e.g., construction zones, rain-obscured jaywalking), 2) Annotating fine-grained states (e.g., black ice, hidden signs), and 3) Developing sim-to-real methods with sensor-accurate degradation. These efforts are vital for certifying autonomy in open-world conditions.

\subsection{Multi-Agent Collaborative Perception Datasets} \label{Multi-Agent datasets}

Recent years have seen significant progress in multi-agent perception datasets, particularly in terms of sensor diversity, collaborative capabilities, and the representation of complex traffic. These datasets provide a critical foundation for BEV-based collaborative perception research. Table~\ref{Collaborative Perception Datasets} summarizes representative datasets across V2V, V2I, V2X, I2I, and roadside scenarios, highlighting their design characteristics, sensor modalities, supported tasks, and relevance to safety-critical BEV applications.

{\LARGE
\begin{table*}[h]
\centering
\caption{Summary of Collaborative Perception BEV Datasets. Task abbreviations: 3DOD-3D Object Detection, MOT-Multi-Object Tracking, MAP—Multi-Agent Perception, MTMCT—Multi-Target Multi-Camera Tracking, MV3DR—Multi-View 3D Reconstruction, PQA—Planning QA, PnP—Perception and Prediction, ReID—Re-Identification, UOD—Unsupervised Object Discovery. “–” indicates unspecified fields.}
\label{Collaborative Perception Datasets}
\begin{adjustbox}{width=\textwidth}
\begin{tabular}{clcccccccccc} 
\toprule
\multirow{2}{*}{\textbf{Type}} & \multirow{2}{*}{\textbf{Datasets}}  & \multirow{2}{*}{\textbf{Year}} & \multirow{2}{*}{\textbf{Venue}} & \multirow{2}{*}{\textbf{Source}} & \multirow{2}{*}{\textbf{Sensors}} & \multirow{2}{*}{\textbf{Tasks}} & \multicolumn{3}{c}{\textbf{Size}}               & \multirow{2}{*}{\textbf{Agents}}  & \multirow{2}{*}{\textbf{Link}}                   \\ 
\cmidrule(lr){8-10}
 &  &   &    &    &    &    & \rule{0pt}{1.0em}\textbf{Image}  & \textbf{LiDAR}  & \textbf{3D BOX}    & \\ 
\midrule
\multirow{13}{*}{I}       & Ko-PER \cite{strigel2014ko} & 2014 & ITSC  & Real  & C\&L  & 3DOD,MOT & 18.7k & 4.8k  & -  & 1  & -\\
                          & CityFlow \cite{tang2019cityflow} & 2019 & CVPR  & Real & C & MTMCT,ReID & 118k   & -  & 22.9k(2D) & 1   &\href{https://cityflow-project.github.io/}{\faLink}  \\
                          & INTERACTION \cite{zhan2019interaction}& 2019  & IROS  & Real  & C\&L & 2DOD,TP  & 1.4M   & -  & 1.4M  & 1 &\href{https://interaction-dataset.com/}{\faLink}   \\
                          & CoopInf \cite{arnold2020cooperative} & 2020 & TITS  & Sim & C & 3DOD  & 10k & -  & 121.2K  & 1 & \href{https://github.com/eduardohenriquearnold/coop-3dod-infra}{\faGithub}   \\
                          & IPS300+ \cite{wang2022ips300+} & 2022 & ICRA & Real & C\&L & 2D/3DOD  & 56.7k  & 14.2k  & 4.5M  & 1 & \href{http://www.openmpd.com/column/IPS300}{\faLink}  \\
                          & LUMPI \cite{busch2022lumpi} & 2022  & IV & Real & C\&L & 3DOD & 200k & 90k   & - & 1 & \href{https://data.uni-hannover.de/cs_CZ/dataset/lumpi}{\faLink} \\
                          & A9-Dataset \cite{cress2022a9} & 2022  & IV  & Real & C\&L & 3DOD  & 1.1k & 1.1k & 14.4k & 1  & \href{https://innovation-mobility.com/en/project-providentia/a9-dataset/}{\faLink}  \\
                          
                          & Rope3D \cite{ye2022rope3d} & 2022 & CVPR  & Real & C\&L & 2D/3DOD & 50k  & - & 1.5M & 12 & \href{https://thudair.baai.ac.cn/rope}{\faLink} \\
                          
                          & TUMTraf-I \cite{zimmer2023tumtraf} & 2023  & ITSC & Real  & C\&L & 3DOD & 4.8k & 4.8k & 57.4k & 1  & \href{https://innovation-mobility.com/en/project-providentia/a9-dataset/}{\faLink}  \\
                          & RoScenes \cite{zhu2024roscenes} & 2024 & ECCV  & Real  & C  & 3DOD  & 1.3M  & -  & 21.13M & 1  & \href{https://roscenes.github.io.}{\faLink}  \\
                          & H-V2X \cite{liu2024h} & 2024 & ECCV & Real & C\&R & BEV Det,MOT,TP & 1.94M  & - & -  & 1 & \href{https://pan.quark.cn/s/86d19da10d18}{\faLink} \\ 
\midrule
\multirow{2}{*}{I2I}      & Rcooper \cite{hao2024rcooper} & 2024 & CVPR & Real & C\&L & 3DOD,MOT & 50k & 30k  & 30k  & 2   & \href{https://github.com/AIR-THU/DAIR-Rcooper}{\faGithub}\\
                          & InScope \cite{zhang2024inscope} & 2024  & arxiv & Real  & L  & 3DOD,MOT  & - & 21.3k  & 188k  & 2  & \href{https://github.com/xf-zh/InScope}{\faGithub}\\ 
\midrule
\multirow{13}{*}{V2V}     & COMAP \cite{yuan2021comap}  & 2021  & ISPRS & Sim & C\&L & 3DOD,SS  & 8.6k & 8.6k & 226.9k  & 2-20  & \href{https://demuc.de/colmap/} 
                          {\faLink}  \\
                          & CODD \cite{arnold2021fast} & 2021  & RA-L & Sim  & L & 3DOD,MOT,SLAM  & -  & 13.5k  & 204k & 4-16  & \href{https://github.com/eduardohenriquearnold/fastreg}{\faGithub} \\
                          & OPV2V \cite{xu2022opv2v} & 2022 & ICRA  & Sim & C\&L\&R  & 3DOD,MOT,SS  & 44k & 11.4k  & 232.9k  & 2-7  & \href{https://mobility-lab.seas.ucla.edu/opv2v/}{\faLink} \\
                          & OPV2V+ \cite{hu2023collaboration} & 2023  & CVPR   & Sim  & C\&L\&R  & 3DOD  & 11.4k+ & 11.4k+ & 232.9k+ & 10  & \href{https://siheng-chen.github.io/dataset/CoPerception+/}{\faLink} \\
                          & V2V4Real \cite{xu2023v2v4real}  & 2023  & CVPR & Real  & C\&L & 3DOD,MOT,S2RDA & 40k & 20k  & 240k  & 2 & \href{https://mobility-lab.seas.ucla.edu/v2v4real/}{\faLink}  \\
                          & LUCOOP \cite{axmann2023lucoop} & 2023  & IV  & Real & L & 3DOD & - & 54k & 7k & 3  & \href{https://data.uni-hannover.de/zh_Hans_CN/dataset/lucoop-leibniz-university-cooperative-perception-and-urban-navigation-dataset}{\faLink} \\
                          & OPV2V-H \cite{lu2024extensible} & 2024 & ICLR  & Sim & C\&L\&R  & 3DOD  & 79k & 79k & 232.9k+ & 2-7 & \href{https://github.com/yifanlu0227/HEAL}{\faGithub} \\
                          & MARS \cite{li2024multiagent} & 2024  & CVPR & Real & C\&L & MAP,MV3DR,UOD  & 15k  & 15k  & - & 2-3  & \href{https://ai4ce.github.io/MARS/}{\faLink}  \\
                          & V2V-QA \cite{chiu2025v2v} & 2025  & arXiv  & Real  & C\&L & 3DOD,PQA  & -  & 18k  & -  & 2   & \href{https://eddyhkchiu.github.io/v2vllm.github.io/}{\faLink}  \\ 
\midrule
\multirow{8}{*}{V2I}       & DAIR-V2X-C \cite{yu2022dair} & 2022  & CVPR & Real & C\&L  & 3DOD & 39k & 39k  & 464k  & 2 & \href{https://air.tsinghua.edu.cn/DAIR-                               V2X/index.html}{\faLink} \\
                          & V2X-Seq \cite{yu2023v2x} & 2023  & CVPR & Real & C\&L & 3DOD,MOT,TP & 71k & 15k & 464k & 2   & \href{https://github.com/AIR-THU/DAIR-V2X-Seq}{\faGithub} \\
                          & HoloVIC \cite{ma2024holovic} & 2024  & CVPR & Real  & C\&L & 3DOD,MOT  & 100k & 100k & 11.47M & 10   & \href{https://holovic.net}{\faGithub} \\
                          & TUMTraf V2X \cite{zimmer2024tumtraf} & 2024 & CVPR & Real  & C\&L & 3DOD,MOT   & 5k  & 2k  & 30k & 2 & \href{https://tum-traffic-dataset.github.io/tumtraf-v2x/}{\faLink}  \\
                          & OTVIC \cite{zhu2024otvic} & 2024  & IROS & Real & C\&L & 3DOD  & 15k & 15k & 24.4k  & 2+ & \href{https://sites.google.com/view/otvic}{\faLink} \\
                          & DAIR-V2XReid \cite{wang2024dair} & 2024  & TITS  & Real  & C\&L & 3DOD, ReID & 2.5k   & -  & -  & 2 & \href{https://github.com/Niuyaqing/DAIR-V2XReid}{\faGithub} \\                        
                          & V2X-Radar \cite{yang2024v2x} & 2024  & arxiv  & Real  & C\&L\&R & 3DOD  & 40k & 20k  & 350k  & 2  & \href{http://openmpd.com/column/V2X-Radar}{\faLink}  \\ 

\midrule
\multirow{13}{*}{V2X}     & V2X-Sim 2.0 \cite{li2022v2x}  & 2022  & RA-L & Sim & C\&L & 3DOD,MOT,SS & 60k & 10K   & 26.6K & 2-5 & \href{https://ai4ce.github.io/V2X-Sim/}{\faLink} \\
                          & DOLPHINS \cite{mao2022dolphins}  & 2022  & ACCV  & Sim & C\&L & 2D/3DOD  & 42.3k & 42.3k  & 292.5k & 3  & \href{https://dolphins-dataset.net/}{\faLink} \\
                          & V2XSet \cite{xu2022v2x} & 2022  & ECCV & Sim  & C\&L & 3DOD  & 44k & 11.4k  & 233k & 2-7  & \href{https://paperswithcode.com/dataset/v2xset}{\faLink}  \\
                          & V2X-Traj \cite{ruan2025learning} & 2024  & NIPS  & Real & C\&L & MP  & 808k & 808k & 1.4M  & 2 & \href{https://github.com/AIR-THU/V2X-Graph}{\faGithub} \\
                           & V2XPnP \cite{zhou2024v2xpnp} & 2024  & arxiv  & Real & C\&L & PnP,TP  & 208k & 40k & 1.45M  & 4  & \href{https://mobility-lab.seas.ucla.edu/v2xpnp/}{\faLink}  \\
                          
                          & DeepAccident \cite{wang2024deepaccident} & 2024 & AAAI & Sim  & C\&L & 3DOD,MOT,SS,MP  & - & 57k & 285k & 1-5 & \href{https://deepaccident.github.io/}{\faLink} \\
                          & V2X-Real \cite{xiang2025v2x}  & 2024 & ECCV  & Real & C\&L & 3DOD & 171k & 33k  & 1.2M  & 4   & \href{https://mobility-lab.seas.ucla.edu/v2x-real}{\faLink}  \\
                          & Multi-V2X \cite{li2024multi}  & 2024  & arxiv & Sim  & C\&L & 3DOD,MOT  & 549k & 146k  & 4.2M   & 0-31 & \href{http://github.com/RadetzkyLi/Multi-V2X}{\faGithub}  \\
                          & Adver-City \cite{karvat2024adver}  & 2024  & arxiv  & Sim & C\&L & 3DOD,MOT,SS & 24k & 24k & 890k & 5 & \href{https://labs.cs.queensu.ca/quarrg/datasets/adver-city/}{\faLink} \\
                          
                          & WHALES \cite{chen2024whales}  & 2024 & arxiv  & Sim & C\&L & 3DOD  & 70k & 17k  & 2.01M  & 8  & \href{https://github.com/chensiweiTHU/WHALES}{\faGithub}  \\
                          & V2X-R \cite{huang2024v2x} & 2024 & arxiv & Sim  & C\&L\&R  & 3DOD  & 150.9k & 37.7k  & 170.8k & 2-7  & \href{https://github.com/ylwhxht/V2X-R}{\faGithub}  \\
                         
                          & SCOPE \cite{gamerdinger2024scope} & 2024 & arxiv & Sim & C\&L & 2/3DOD,SS,S2RDA  & 17k  & 17k  & 575k & 3-21 & \href{https://ekut-es.github.io/scope/}{\faLink}\\
                          & Mixed Signals \cite{luo2025mixed} & 2025  & arxiv  & Real & L  & 3DOD & - & 45.1k & 240.6k & 4 & \href{https://mixedsignalsdataset.cs.cornell.edu/}{\faLink}\\
\bottomrule
\end{tabular}
\end{adjustbox}
% \end{tabular}
\end{table*}
}

\subsubsection{Infrastructure-Side Perception Datasets}
Infrastructure-side perception leverages fixed roadside units (RSUs) to enable high-precision, wide-area environmental sensing. Compared to vehicle-mounted systems, RSUs provide elevated and stable viewpoints, mitigating occlusion and eliminating ego-motion artifacts. As a complementary modality, they extend perception range, support multi-agent collaboration, and strengthen safety-critical BEV tasks. This subsection reviews representative infrastructure-side datasets and their roles in promoting safety, robustness, and coordination, categorized into standalone roadside and I2I configurations.

\paragraph{Roadside-only Perception Datasets} Roadside perception, enabled by fixed RSUs on poles or gantries, provides a stable and occlusion-resilient view with broad spatial and temporal coverage. Compared to ego-vehicle sensors, its elevated and static placement supports reliable monitoring of intersections, congestion, and long-range targets. As an independent yet complementary sensing modality, it plays a key role in enhancing the safety, robustness, and collaboration of BEV-based autonomous driving systems.

Early datasets such as Ko-PER~\cite{strigel2014ko} and CityFlow~\cite{tang2019cityflow} focused on object re-identification but were limited in scale and annotation quality. IPS300+\cite{wang2022ips300+} improves spatial coverage at intersections, while LUMPI\cite{busch2022lumpi} and TUMTraf-I~\cite{zimmer2023tumtraf} add multi-sensor and temporal diversity. For highways, TUMTraf-A9~\cite{cress2022a9} captures varied road types and weather. Rope3D~\cite{ye2022rope3d} supports monocular 3D detection, RoScenes~\cite{zhu2024roscenes} targets congestion modeling, and H-V2X~\cite{liu2024h} integrates radar-vision fusion for enhanced resilience.

Despite these advances, existing roadside datasets still lack comprehensive coverage of extreme weather events, rare safety-critical incidents, and detailed occlusion annotations. Moreover, their static deployment and limited spatial diversity hinder generalization to broader scenarios. Future datasets should emphasize dynamic environments, heterogeneous sensors, and deployment scalability to support robust infrastructure-side BEV perception fully.

\paragraph{I2I Perception Datasets}

I2I perception datasets enable collaborative sensing across multiple RSUs, thereby overcoming the limitations of single-RSU systems in spatial coverage, occlusion mitigation, and continuity. By fusing data from distributed and heterogeneous nodes, these datasets support robust environmental modeling in dense urban settings and extended road networks, aligning well with BEV-based collaborative perception.

Representative efforts include RCooperr~\cite{hao2024rcooper}, which provides large-scale real-world data across intersections and corridors with diverse LiDAR-camera configurations for 3D detection and tracking. InScope~\cite{zhang2024inscope} emphasizes occlusion-aware perception via multi-LiDAR fusion and introduces evaluation metrics for blind-spot coverage under constrained views. These datasets highlight the potential of I2I configurations in enhancing perception resilience under complex layouts, occlusions, and multi-agent interactions.

However, existing I2I datasets still struggle with cross-RSU calibration, temporal synchronization, and scalability in deployment. Moreover, most fail to capture long-term dynamics or rare-event scenarios. Future work should focus on standardization, heterogeneous sensor integration, and deployment diversity to enable scalable, robust BEV perception in intelligent infrastructure networks.

\subsubsection{V2V Perception Datasets}

V2V collaborative perception enhances environmental awareness through real-time information exchange among multiple vehicles, extending perception range and improving robustness in complex scenarios. Recent V2V datasets support BEV-centric collaboration by enabling synchronized multi-agent data collection, multimodal sensing, and diverse traffic conditions.

Recent V2V datasets address the spatial and temporal limitations of single-agent systems by incorporating more agents, enhanced sensor fusion, and realistic environments. For instance,
OPV2V~\cite{xu2022opv2v} established benchmarks for BEV feature fusion under ideal synchronization, followed by OPV2V+\cite{hu2023collaboration} with increased agent diversity and OPV2V-H\cite{lu2024extensible} incorporating heterogeneous LiDAR-camera fusion. V2V4Real~\cite{xu2023v2v4real} bridges simulation and real-world data, supporting 3D detection and tracking. LUCOOP~\cite{axmann2023lucoop} and MARS~\cite{li2024multiagent} offer multi-vehicle recordings across varied environments, with MARS focusing on adverse weather and dense traffic. V2V-QA~\cite{chiu2025v2v} explores large language model integration for joint reasoning and planning.

While these datasets advance occlusion handling, long-range detection, and robustness, gaps remain. Most lack fine-grained annotations for rare events, occluded agents, and near-miss incidents. Short sequence lengths hinder long-term reasoning, while factors such as calibration errors, limited radar coverage, and the absence of high-definition maps further constrain their applicability in real-world scenarios.

\subsubsection{V2I Perception Datasets}

V2I collaborative perception integrates data from RSUs and vehicle-side sensors to improve robustness and accuracy under occlusion, high-speed traffic, and low-light conditions. Recent V2I datasets support BEV-based multi-agent perception in dynamic and safety-critical environments.

V2I datasets have evolved significantly in scale, detail, and environmental diversity to meet the demands of intelligent transportation systems.DAIR-V2X-C~\cite{yu2022dair} offers high-quality multimodal data for urban BEV perception, while V2X-Seq~\cite{yu2023v2x} includes rich trajectory annotations for forecasting and risk assessment. HoloVIC~\cite{ma2024holovic} and TUMTraf-V2X~\cite{zimmer2024tumtraf} target occlusion-prone intersections, capturing complex interactions and high-resolution scenes. OTVIC~\cite{zhu2024otvic} provides multi-view, multimodal data under high-speed and noisy conditions. V2X-Radar~\cite{yang2024v2x} enhances performance in adverse weather via radar integration.

These datasets significantly advance BEV perception in safety-critical V2I scenarios, especially in conditions involving occlusions and dense traffic. They enable fine-grained tasks such as trajectory prediction and occlusion-aware detection with broad coverage and high spatial precision. However, limitations persist in geographic diversity, environmental variability, and RSU heterogeneity.

\subsubsection{V2X Perception Datasets}

V2X perception integrates V2V and V2I collaboration by fusing multimodal sensor data from vehicles and infrastructure, enabling extended perception range, temporal consistency, and enhanced scene understanding for robust BEV-based autonomous driving.

Recent V2X datasets have evolved to support diverse agent configurations, complex environmental conditions, and advanced multi-task learning. V2X-Sim 2.0~\cite{li2022v2x} pioneered multi-task simulation for collaborative perception. DOLPHINS~\cite{mao2022dolphins}, V2X-Traj~\cite{ruan2025learning} and V2XPnP~\cite{zhou2024v2xpnp} model multi-agent interaction with sequential dynamics. To address adverse conditions, V2XSet~\cite{xu2022v2x} and SCOPE~\cite{gamerdinger2024scope} simulate real-world noise and uncertainty. DeepAccident~\cite{wang2024deepaccident}, Adver-City~\cite{karvat2024adver}, and V2X-R~\cite{huang2024v2x} focus on robustness in extreme weather and safety-critical events.  V2X-Real~\cite{xiang2025v2x}, Multi-V2X~\cite{li2024multi}, and WHALES~\cite{chen2024whales}expand scalability through heterogeneous collaboration and diverse scenarios, supporting 3D annotations and dynamic scheduling.

These datasets facilitate perception fusion, interaction modeling, and all-weather sensing in complex scenarios. Yet, challenges persist in annotation costs, geographic diversity, and coverage of rare events. Future progress requires hybrid real-simulated datasets, broader environmental variety, and stronger generalization for scalable, real-world V2X deployment.

\subsection{Safety and Robustness in Datasets} \label{Security and Robustness Analysis}
\begin{figure*}[t]
    \centering
    \includegraphics[width=\textwidth]{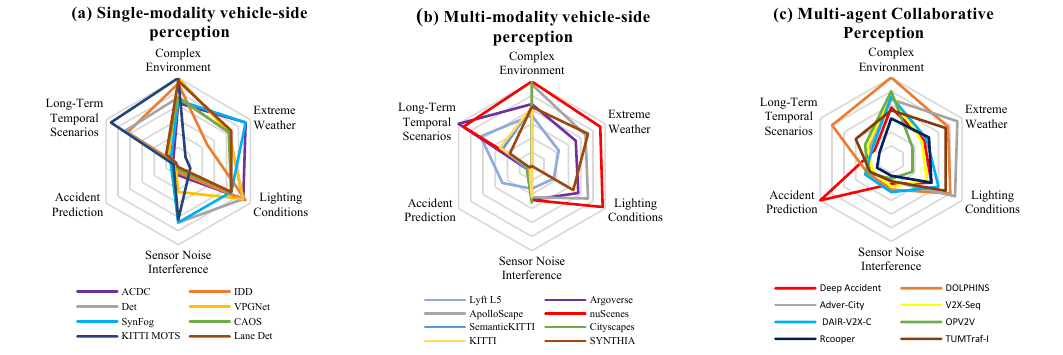}
    \caption{Comparative analysis of dataset support across safety-critical scenarios. The radar charts illustrate the capabilities of BEV perception datasets in addressing six key challenges: complex environments, extreme weather, varying lighting conditions, sensor noise interference, accident prediction, and long-term temporal reasoning. Subfigures correspond to (a) single-modality vehicle-side datasets (\textit{SafeBEV 1.0}), (b) multimodality vehicle-side datasets (\textit{SafeBEV 2.0}), and (c) multi-agent collaborative datasets (\textit{SafeBEV 3.0)}.}
    \label{lidar_dataset}
\end{figure*}
Safety and robustness are essential prerequisites for reliable autonomous driving, particularly in complex and uncertain environments. As the foundation of perception system development and benchmarking, datasets must effectively support these critical capabilities. To this end, we assess BEV perception datasets from a safety-oriented perspective, focusing on their ability to represent diverse and challenging real-world conditions. Fig.~\ref{lidar_dataset} summarizes the key scenario coverage across datasets. This section categorizes and analyzes them across three progressive stages: SafeBEV 1.0 (single-modality), SafeBEV 2.0 (multimodality), and SafeBEV 3.0 (multi-agent collaboration).

\subsubsection{SafeBEV 1.0 Datasets}

Single-modality vehicle-side datasets, built upon individual sensors such as cameras or LiDAR, serve as foundational benchmarks for early BEV perception research under constrained sensing conditions. These datasets often target isolated tasks with relatively low complexity but limited environmental diversity. To improve robustness against adverse conditions, Foggy-C~\cite{foggy-cityscape}, ACDC~\cite{ACDC}, and SynFog~\cite{synfog} incorporate synthetic weather variations to simulate fog, rain, and low-light environments. For dynamic and complex scenes, KITTI MOTS~\cite{Mots} and IDD~\cite{IDD} contribute with motion segmentation and diverse traffic patterns. However, these datasets still fall short in supporting safety-critical scenarios, such as accident prediction, long-term temporal reasoning, sensor degradation, and realistic lighting conditions. Furthermore, the domain shift between synthetic and real-world data remains a barrier to generalizability.

\subsubsection{SafeBEV 2.0 Datasets}

Multimodality vehicle-side datasets integrate heterogeneous sensing modalities to capture the driving environment from diverse perspectives, thereby strengthening system robustness and enabling safety-aware autonomous perception. Representative datasets like Cityscapes~\cite{cityscapes}, ApolloScape~\cite{apolloscape}, and nuScenes~\cite{nuscenes} emphasize dense urban scenes and diverse traffic contexts, supporting robust perception in complex environments. For accident prediction, Lyft L5~\cite{Lyft} provides temporally rich sequences that facilitate early risk detection. Additionally, Argoverse~\cite{Argoverse}, SYNTHIA~\cite{synthia}, and nuScenes offer extensive coverage of adverse weather, varying illumination, and sensor degradation, reinforcing robustness under real-world uncertainties. Despite these strengths, most datasets still offer limited support for high-risk scenarios such as collision-prone interactions and extreme environmental conditions, indicating room for improvement in safety-critical context modeling.

\subsubsection{SafeBEV 3.0 Datasets}

Multi-agent collaborative datasets enhance single-vehicle systems by enabling broader perception coverage and greater resilience in complex traffic environments. Through inter-agent collaboration, they support occlusion-aware perception, trajectory forecasting, and long-range safety reasoning—key components of robust BEV perception. Simulated datasets such as OPV2V~\cite{xu2022opv2v}, DOLPHINS~\cite{mao2022dolphins}, and Adver-City~\cite{karvat2024adver} address occlusions, long-term dynamics, and adverse weather, while DeepAccident~\cite{wang2024deepaccident} focuses on rare collisions for safety-critical learning. Real-world datasets such as DAIR-V2X-C~\cite{yu2022dair}, V2X-Traj~\cite{ruan2025learning}, and RCooper~\cite{hao2024rcooper} address challenges in dense urban environments, blind-spot mitigation, and infrastructure-based collaboration through the use of heterogeneous sensors. However, limitations persist in cross-agent calibration, semantic alignment, and rare-event coverage, highlighting the need for more scalable and standardized real-world benchmarks.

\subsubsection{Limitations and Future Work}

Despite progress from single-modality systems to multi-agent collaboration, current BEV datasets still face key limitations. Safety-critical scenarios—such as rare accidents, occlusions, and corner cases—remain underrepresented, limiting the evaluation of real-world robustness. Temporal discontinuities and insufficient modeling of sensor degradation further hinder predictive capabilities and uncertainty-aware perception. While datasets like SafeBEV 3.0 introduce spatial diversity and multi-agent collaboration, they often lack standardized calibration, precise synchronization, and scalability across regions and infrastructures. Advancing safe autonomous driving requires moving beyond static annotations and toward dynamic scene evolution, causal interaction modeling, and the quantification of uncertainty. Bridging the sim-to-real gap, enabling large-scale coordination, and aligning dataset design with real-world risk distributions are essential for achieving robust BEV perception in open-world conditions.

{\LARGE
\begin{table*}[t]
\centering
% \caption{Robustness and Safety Evaluation of BEV Perception Algorithms under Extreme Conditions. Results are reported on two standardized benchmarks: nuScenes-C~\cite{xie2025benchmarking} and BEV-Robust~\cite{zhu2023understanding}. nuScenes-C evaluates performance degradation under sensor corruptions including camera failure, quantization, motion blur, and environmental distortions. BEV-Robust assesses robustness under noise, blur, digital artifacts, and weather variations. ``–'' indicates unavailable results. All values are percentages (\%).}
\caption{Evaluation of the robustness and safety of BEV perception algorithms under extreme conditions. Results are reported on two standardized benchmarks: nuScenes-C, which assesses performance degradation under camera failure, quantization, motion blur, and environmental settings; and BEV-Robust, which evaluates robustness against noise, blur, digital artifacts, weather conditions, and normal scenes. Metrics include mCE (mean Corruption Error), mRR (mean Relative Robustness), mAP (mean Average Precision), and NDS (nuScenes Detection Score). All values are shown in percentage (\%). ``–'' indicates unavailable results.}
\label{Robustness Evaluation}
\begin{adjustbox}{width=\textwidth}
\begin{tblr}{
  colspec = {Q[l,1.4cm] Q[l,2.6cm] Q[c,1.2cm] Q[c,1.2cm] Q[c,1.2cm] Q[c,1.5cm] | Q[c,1.1cm] Q[c,1.1cm] Q[c,1.1cm] Q[c,1.1cm] Q[c,1.1cm]},
  row{2} = {c},
  cell{1}{1} = {r=2}{},
  cell{1}{2} = {r=2}{},
  cell{1}{3} = {c=4}{c},
  cell{1}{7} = {c=5}{c},
  cell{3}{1} = {r=5}{c},
  cell{6}{7} = {c},
  cell{6}{8} = {c},
  cell{6}{9} = {c},
  cell{6}{10} = {c},
  cell{6}{11} = {c},
  cell{7}{7} = {c},
  cell{7}{8} = {c},
  cell{7}{9} = {c},
  cell{7}{10} = {c},
  cell{7}{11} = {c},
  cell{8}{1} = {r=4}{c},
  cell{9}{3} = {c},
  cell{9}{4} = {c},
  cell{9}{5} = {c},
  cell{9}{6} = {c},
  cell{10}{7} = {c},
  cell{10}{8} = {c},
  cell{10}{9} = {c},
  cell{10}{10} = {c},
  cell{10}{11} = {c},
  cell{11}{3} = {c},
  cell{11}{4} = {c},
  cell{11}{5} = {c},
  cell{11}{6} = {c},
  vline{4} = {1}{0.05em},
  vline{7} = {1-11}{0.05em},
  hline{1,12} = {-}{0.08em},
  hline{2} = {3-11}{0.03em},
  hline{3,8} = {-}{0.05em},
}
    Category      & Method      & nuScenes-C \cite{xie2025benchmarking} ($\mathrm{mCE}\downarrow\!/\;\mathrm{mRR}\uparrow$)  &              &              &               & BEV-Robust \cite{zhu2023understanding} ($\mathrm{mAP}\uparrow\!/\;\mathrm{NDS}\uparrow$) &            &            &           &           \\
           &             & Camera              & Quant        & Motion       & Setting & Noise~              & Blur       & Digital    & Weather   & Normal    \\
           
SafeBEV1.0 & DETR3D \cite{wang2022detr3d}     & 100.0/64.7        & 100.0/75.2 & 100.0/63.0 & 100.0/74.7  & 16.3/28.1           & 14.5/25.0  & 29.9/38.7  & 24.5/32.6 & 34.7/42.2 \\
           & BEVDepth \cite{li2023bevdepth}   & 104.7/58.9        & 106.2/67.8 & 102.0/61.9 & 115.6/51.8  & 5.1/12.7            & 16.1/25.4  & 25.5/34.2  & 26.5/35.7 & 33.2/40.4 \\
           & BEVDet \cite{huang2021bevdet}     & 107.2/58.5        & 111.3/63.9 & 108.2/54.7 & 121.8/44.8  & 3.3/10.5            & 11.6/20.65 & 21.22/30.5 & 23.8/33.8 & 29.2/37.2 \\
           & PETR \cite{liu2022petr}       & 106.7/61.2        & 110.3/67.5 & 104.9/62.7 & 114.8/59.4  & -                   & -          & -          & -         & -         \\
           & PolarFormer \cite{jiang2023polarformer} & 96.7/64.6          & 95.1/76.3  & 92.7/70.0  & 96.9/73.9   & -                   & -          & -          & -         & -         \\
SafeBEV2.0 & BEVFormer \cite{li2022bevformer}  & 93.2/59.6          & 101.2/67.8 & 99.5/52.1~ & 99.7/61.0   & 17.2/33.1           & 16.9/31.1  & 32.8/45.2  & 31.7/41.7 & 37.0/47.9 \\
           & BEVFusion \cite{liu2023bevfusion}  & -                   & -            & -            & -             & 63.4/68.7           & 65.3/69.7  & 67.2/70.7  & 60.6/63.4 & 68.5/71.4 \\
           & BEVerse \cite{zhang2022beverse}    & 94.8/66.7         & 108.5/55.7 & 100.2/56.7 & 121.8/35.8  & -                   & -          & -          & -         & -         \\
           & TransFusion \cite{bai2022transfusion} & -                   & -            & -            & -             & 65.9/70.2           & 66.5/70.4  & 66.9/70.7  & 58.1/62.4 & 67.2/70.9 
\end{tblr}
\end{adjustbox}
\end{table*}
}

\subsection{Robustness and Safety Benchmarking on BEV Datasets}  \label{Metrics and Benchmark}

The safety and reliability of autonomous driving hinge on the robustness of perception algorithms under challenging conditions, including sensor failures, adverse weather, and poor lighting. This section evaluates representative BEV models under various corruptions, following standardized protocols from nuScenes-C~\cite{xie2025benchmarking} and BEV-Robust~\cite{zhu2023understanding}. Table~\ref{Robustness Evaluation} summarizes model performance across different corruption types, highlighting their resilience and failure patterns in safety-critical scenarios.

\subsubsection{Evaluation Metrics}

To evaluate BEV model performance under extreme conditions, we adopt four representative metrics. Mean Average Precision (mAP) measures detection accuracy, while NuScenes Detection Score (NDS)\cite{nuscenes} extends mAP by integrating five true-positive metrics for a more comprehensive assessment of 3D detection and semantic consistency. Mean Corruption Error (mCE)\cite{xie2025benchmarking} quantifies robustness by comparing a model’s NDS under corruption to a baseline (DETR3D), with lower values indicating better performance. Mean Relative Robustness (mRR)~\cite{xie2025benchmarking} measures performance retention relative to clean inputs, where higher values denote greater resilience. Together, mAP and NDS assess accuracy, while mCE and mRR reflect robustness under adverse conditions.

\subsubsection{Benchmark-Based Robustness Analysis}
Benchmark results reveal that BEV perception algorithms—regardless of modality—experience substantial performance degradation under corrupted and extreme conditions. Single-modality models such as BEVDet and BEVDepth are particularly susceptible: on BEV-Robust, BEVDet’s mAP drops to 3.3\% under noise and 11.6\% under blur, with NDS falling below 11\%. BEVDepth performs similarly poorly, reaching just 5.1\% mAP and 12.7\% NDS under the same corruptions. On nuScenes-C, models like BEVDet and BEVerse report mCE values exceeding 120\%, with mRR declines of more than 50\%, highlighting their limited robustness in adverse environments.

In contrast, multimodal fusion methods demonstrate significantly improved resilience. BEVFusion and TransFusion consistently maintain mAP and NDS above 60\% across various corruptions, showing less than 10\% performance drop from clean conditions. BEVFormer also performs robustly under nuScenes-C corruptions, maintaining mCE below 102\% and mRR above 96\%. These results underscore the advantages of sensor redundancy and effective fusion strategies in mitigating performance collapse.

Despite these gains, all models exhibit non-negligible degradation, underscoring the persistent gap between current BEV systems and safety-critical deployment requirements. These findings call for continued research into corruption-aware training, uncertainty modeling, and benchmarking protocols that better reflect real-world resilience needs.

\subsubsection{Limitations and Future Work}

While current benchmarks offer valuable insights into BEV model robustness, most remain limited to single-vehicle settings and overlook the unique challenges of multi-agent collaborative perception. Existing evaluations cover a narrow range of corruptions and often fail to capture real-world complexities such as sensor misalignment, cross-agent asynchrony, and large-scale occlusions. Moreover, many rely on static or synthetic perturbations, lacking dynamic modeling of temporal evolution, long-tail events, or compound failures. Although multimodal fusion shows potential, its integration strategies often lack robustness under degraded or conflicting sensor inputs.

To address these gaps, future efforts should develop unified, corruption-aware evaluation protocols that generalize across diverse sensors and agent interactions. This includes adversarial or simulation-augmented benchmarks reflecting real-world risks, and robustness metrics beyond mAP or NDS, such as temporal stability, uncertainty calibration, and task-level safety indicators. Scalable benchmarks that incorporate V2X communication noise and heterogeneous agent behavior are essential for advancing system-level resilience in safety-critical environments.

\section{Challenges} \label{sec6}
In this section, we highlight four key challenges hindering the real-world deployment of BEV perception in autonomous driving. Section~\ref{unkown_environment} examines generalization to out-of-distribution object categories under open-world scenarios. Section~\ref{large_data} discusses the limited availability of labeled data for multi-agent learning. Section~\ref{undefined_sensor} investigates the impact of sensor uncertainty on system robustness. Section~\ref{Multi-agent communication} analyzes how communication delays impair collaborative perception performance.

\subsection{Generalization to Open-World Object Categories}
\label{unkown_environment}

Most BEV perception systems are developed under the closed-set assumption, where object categories are fixed and known a priori. However, real-world environments are inherently open, requiring autonomous vehicles to interact with out-of-distribution object categories not encountered during training. This discrepancy undermines the robustness and safety of current BEV systems. A central challenge lies in achieving category-agnostic perception while preserving spatial precision, particularly in cluttered or dynamic scenes. Open-world perception introduces new problems: (1) distinguishing novel from known categories; (2) generalizing to unseen semantics without explicit annotations; and (3) avoiding catastrophic forgetting when incrementally learning new categories.

Recent efforts focus on generalizable and semantically rich frameworks. Camera-based occupancy networks~\cite{mescheder2019occupancy,roddick2020predicting} model voxelized occupancy rather than bounding boxes, enabling class-agnostic spatial reasoning. Vision-language models (VLMs) like CLIP~\cite{radford2021learning} and SEEM~\cite{zou2023segment} employ large-scale pretraining and language prompts for zero-shot or few-shot detection. Approaches such as YOLO-World~\cite{cheng2024yolo}, YOLO-UniOW~\cite{liu2024yolo}, and YOLOE~\cite{wang2025yoloe} further extend open-vocabulary capabilities through VLM integration. Foundation models including SAM~\cite{kirillov2023segment} and DINO~\cite{oquab2023dinov2,caron2021emerging} also contribute strong visual priors for open-world BEV perception.

\subsection{Insufficient Labeled Data for Multi-Agent Perception}
\label{large_data}

Multi-Agent Collaborative Perception (MACP) enhances situational awareness by fusing data from spatially distributed vehicles. While data collection is feasible, annotation poses the main challenge. Unlike single-agent datasets, MACP requires synchronized multi-view sequences with detailed labels such as masks and trajectories, demanding precise spatio-temporal alignment. Additionally, varied sensor setups, communication delays, and dynamic conditions further increase labeling complexity and hinder scalability.

To mitigate this, recent efforts explore: (1) \textbf{Self-supervised learning}, using contrastive objectives to extract features without labels~\cite{gosala2023skyeye,sun2023calico}; (2) \textbf{Pseudo-labeling}, where teacher-student models leverage predictions from foundation models such as SAM~\cite{kirillov2023segment}, CLIP~\cite{radford2021learning}, and DINO~\cite{oquab2023dinov2}; (3) \textbf{Semi-supervised learning}, combining labeled and unlabeled data through consistency-based training; and (4) \textbf{Synthetic data generation}, using simulators like CARLA or world models~\cite{hu2024drivingworld,peng2024towards} to produce scalable multi-agent datasets. While promising, these methods still face challenges in label quality, cross-agent alignment, and domain generalization for real-world applications.

\subsection{Robustness under Sensor-Induced Uncertainty}
\label{undefined_sensor}

In multi-sensor fusion BEV systems, sensor uncertainty and failure remain core challenges. Adverse conditions—such as rain, snow, fog, or low light—can degrade sensor input quality, introducing noise, missing data, or artifacts. Hardware faults (e.g., occlusion, malfunction, or link failure) and temporal drift (e.g., calibration error or sensitivity shifts) further increase anomaly risks. As sensor count and diversity grow, the probability of at least one failure—given by $1-(1-p)^n$ for per-sensor failure rate $p$—rises sharply, often far exceeding $p$, thereby elevating system-level safety risks. These issues make maintaining robust BEV perception difficult in real-world deployments.

Recent advances have sought to address these challenges through uncertainty modeling and robust training strategies. Bayesian neural networks~\cite{kendall2017uncertainties} enable explicit quantification of epistemic and aleatoric uncertainty, while data augmentation methods such as AugMix~\cite{hendrycks2019augmix}, CutMix~\cite{yun2019cutmix}, and point cloud sparsification enhance resilience to sensor degradation. More recently, elastic perception frameworks~\cite{han2021dynamic} have been proposed, allowing systems to dynamically adapt inference pathways based on real-time sensor quality assessment. Nevertheless, consistently robust BEV perception under highly variable and unpredictable sensor conditions remains an unresolved challenge, particularly for safety-critical applications.

\subsection{Multi-Agent Communication Delay}
\label{Multi-agent communication}

In multi-agent BEV perception, timely and reliable communication is crucial for fusing spatial and semantic information across distributed vehicles and infrastructure nodes. However, real-world networks suffer from limited bandwidth, synchronization drift, and variable latency~\cite{wang2020v2vnet,yu2022dair,xu2022opv2v}, leading to stale or misaligned features that degrade perception accuracy. This becomes especially problematic in high-speed or congested traffic, where even minor delays may jeopardize safety-critical decisions. Further complications arise from asynchronous sensor sampling, heterogeneous viewpoints, and diverse transmission protocols, which collectively impair temporal alignment~\cite{khani2023collaborative,lu2024extensible}. These inconsistencies challenge the robustness of collaborative perception, particularly when agents must rely on outdated or partial observations.

To address these issues, recent works propose latency-aware pipelines, including feature compression~\cite{li2021learning}, asynchronous fusion~\cite{arnold2021fast}, and predictive perception frameworks~\cite{li2024multi}. Despite these advances, scalable and real-time multi-agent BEV perception under communication uncertainty remains an open and critical challenge for the deployment of safe and robust autonomous systems.

\section{Outlook} \label{sec7}

As BEV perception advances from static scene parsing toward intelligent, adaptive driving, future progress will depend on cognitive reasoning, behavior modeling, and foundation models. This section outlines four emerging directions: Section~\ref{Embodied Agents} explores cognition-driven BEV in embodied agents; Section~\ref{Behavior-Aligned E2E} investigates behavior-aligned end-to-end driving; and Section~\ref{Foundation Models} discusses generalization via foundation models.

\subsection{Cognition-Driven BEV Perception in Embodied Agents} \label{Embodied Agents}

Current BEV perception algorithms fuse multimodal features into spatial-semantic maps but lack adaptability to dynamic environments and active interaction. Embodied intelligence bridges this gap by integrating perception, decision, and action through closed-loop feedback, enabling context-aware behavior in real-world settings~\cite{zhou2025opendrivevla},~\cite{zhao2025cot},~\cite{sapkota2025vision}. Future BEV systems should go beyond static mapping to support seamless perception-action coupling through environmental interaction.

This integration spans three dimensions: (1) Scene Understanding and Prediction: Agents must interpret complex traffic, infer intent, and anticipate changes for proactive adaptation. (2) Perception-Guided Decision-Making: Real-time sensing should drive both reflexive responses and long-term maneuvers like lane changes or overtaking. (3) Collective Interaction: Robust V2X and human-vehicle interaction will support collaborative behaviors through implicit communication, enabling collective intelligence akin to flocking or schooling.

\subsection{Behavior-Aligned End-to-End BEV Driving} \label{Behavior-Aligned E2E}
Conventional modular pipelines separating perception, planning, and control often suffer from inefficiencies and latency in complex scenarios. End-to-end autonomous driving addresses this by directly mapping sensor inputs to planning/control outputs, improving coordination and real-time responsiveness. Recent advances in end-to-end autonomous driving have highlighted the importance of \emph{behavior alignment}, which refers to ensuring that the driving policy not only accomplishes navigation tasks but also exhibits human-like, safe, and socially compliant driving behaviors~\cite{hu2023planning,jiang2023vad,chen2024vadv2}. 

% Most frameworks adopt BEV as the backbone and integrate high-level semantic inputs—such as ego-motion, maps, tasks, or language embeddings—to enhance contextual understanding.

Despite progress, these models face slow convergence, limited behavioral interpretability, and robustness challenges. Incorporating anchor paths (e.g., sparse waypoints, candidate trajectories) as auxiliary supervision helps reduce policy uncertainty and guide decision-making. Closed-loop reinforcement learning further refines policies through real-time feedback, forming an adaptive perception–decision–control loop. To address data scarcity and improve long-tail generalization, generative methods like NeRF~\cite{mildenhall2021nerf} and diffusion models can synthesize diverse, high-fidelity data for scalable training.

\subsection{Generalizable BEV Perception with Foundation Models}  \label{Foundation Models}

Conventional BEV perception models are typically trained on closed-set categories such as pedestrian, car, and bus. However, real-world environments are inherently open-ended, and such rigid label spaces limit the model’s ability to recognize novel objects or unforeseen scenarios, posing safety risks. Foundation models~\cite{liu2023visual},~\cite{radford2021learning},~\cite{brown2020language},~\cite{radford2019language}, pretrained on large-scale data, offer broad visual representations and strong generalization across diverse tasks.

When adapted to autonomous driving, these models~\cite{jiang2024senna},~\cite{huang2024drivemm},~\cite{wang2023drivemlm} demonstrate effective multimodal fusion across cameras, LiDAR, and HD maps for rich semantic understanding. They offer three key benefits: (1) improved recognition of diverse traffic participants and long-tail objects, enhancing safety; (2) enhanced reasoning in complex scenes for trajectory and intent prediction, aiding high-level decision-making; and (3) inherent interpretability through scene-level QA and reasoning, fostering user trust. Their strong transferability enables rapid adaptation to new domains with minimal fine-tuning, accelerating BEV perception deployment.

\section{Conclusion}
In this survey, we present the first systematic and comprehensive review of both vehicle-side and collaborative BEV perception methods, with a particular emphasis on safety and robustness. We categorize existing approaches into three progressive stages—single-modality vehicle-side perception, multi-modality vehicle-side perception, and multi-agent collaborative perception—thereby elucidating the evolution of BEV perception systems and providing a critical assessment of their strengths and limitations across diverse operational scenarios. Additionally, we analyze representative public datasets and establish standardized evaluation protocols and benchmarks, offering valuable resources to guide future research and development. To foster transparency and reproducibility, we release an open-source repository containing method implementations and dataset usage guidelines. Finally, we identify and discuss key challenges, including open-world deployment, large-scale unlabeled data, sensor degradation, and multi-agent communication delay, and explore future directions such as integration with embodied intelligence and end-to-end autonomous driving. We hope this survey can serve as a foundational reference for the community, catalyzing further advances toward safe, robust, and intelligent BEV perception systems.

% \bibliography{sn-bibliography}% common bib file

% \small\bibliography{sn-bibliography}
% \bibliographystyle{ieeetr}    
% \small\bibliography{main}

% \bibliographystyle{sn-basic} % 推荐使用模板自带的样式之一
% \bibliography{bibliography}

% ----------------------
% REFERENCES
% ----------------------
%\bibliographystyle{bibliography} % 也可以是 sn-mathphys 等
\bibliography{bibliography}

\end{document}